\documentclass[twoside]{article}

\usepackage[accepted]{aistats2025}
\usepackage{natbib}

\usepackage[breaklinks,colorlinks]{hyperref}
\usepackage[dvipsnames]{xcolor}
\hypersetup{
   colorlinks,
   linkcolor={red!60!black},
   citecolor={green!40!black}
}

\usepackage{textcomp}
\usepackage[utf8]{inputenc}
\usepackage[T1]{fontenc}
\usepackage{bbm}
\usepackage{amssymb}
\usepackage{amsmath}
\usepackage{amsthm}
\usepackage{mathtools}
\usepackage{algorithm,algorithmic}
\usepackage{stmaryrd}
\usepackage{booktabs}
\usepackage{color}
\usepackage{nicefrac}
\usepackage{multirow}
\usepackage{multicol}
\usepackage{lipsum}  
\usepackage{tikz}
\usepackage{pgfplots}
\usepgfplotslibrary{groupplots}
\usepgfplotslibrary{fillbetween}
\usepackage{adjustbox}
\usepackage{wrapfig}
\usepackage{setspace}
\usepackage[font=small]{caption}

\DeclareMathOperator*{\argmin}{\mathrm{arg\,min}}
\DeclareMathOperator*{\argmax}{\mathrm{arg\,max}}

\newtheorem{theorem}{Theorem}[section]

\newtheorem{remark}[theorem]{Remark}
\newtheorem{definition}[theorem]{Definition}

\newcommand{\q}[1]{``#1''}
\usepackage[capitalize]{cleveref}
\crefname{section}{Sec.}{Secs.}
\Crefname{section}{Section}{Sections}
\Crefname{table}{Table}{Tables}
\crefname{table}{Tab.}{Tabs.}

\usepackage{url}
\usepackage{amsfonts}
\usepackage[nopatch=footnote]{microtype}
\usepackage{xcolor} 
\usepackage{soul}

\definecolor{plotcolor1}{HTML}{e41a1c}
\definecolor{plotcolor2}{HTML}{377eb8}
\definecolor{plotcolor3}{HTML}{4daf4a}
\definecolor{plotcolor4}{HTML}{e7298a}
\definecolor{plotcolor5}{HTML}{ff7f00}
\definecolor{plotcolor6}{HTML}{666655}
\definecolor{plotcolor7}{HTML}{984ea3}

\usetikzlibrary{arrows.meta}
\usetikzlibrary{shapes.misc}

\tikzset{cross/.style={cross out, draw=black, minimum size=2*(#1-\pgflinewidth), inner sep=0pt, outer sep=0pt},
cross/.default={1pt}}

\usepackage{graphicx}
\usepackage{gensymb}
\usepackage{caption}
\usepackage{subcaption}

\usepackage{pgfplotstable}

\begin{document}

%

%
\runningauthor{Sadiku, Wagner, Nagarajan, Pokutta}

\twocolumn[

\aistatstitle{S-CFE: Simple Counterfactual Explanations}

\aistatsauthor{Shpresim Sadiku$^{1,2}$ \And Moritz Wagner$^{1,2}$ \And Sai Ganesh Nagarajan$^1$ \And Sebastian Pokutta$^{1,2}$}
 \vspace{8pt}
  \aistatsaddress{$^{1}$Zuse Institute Berlin \\ $^{2}$Technische Universität Berlin} 
   
]

\begin{abstract}
    We study the problem of finding optimal sparse, manifold-aligned counterfactual explanations for classifiers. Canonically, this can be formulated as an optimization problem with multiple non-convex components, including classifier loss functions and manifold alignment (or \emph{plausibility}) metrics. The added complexity of enforcing \emph{sparsity}, or shorter explanations, complicates the problem further. Existing methods often focus on specific models and plausibility measures, relying on convex $\ell_1$ regularizers to enforce sparsity. In this paper, we tackle the canonical formulation using the accelerated proximal gradient (APG) method, a simple yet efficient first-order procedure capable of handling smooth non-convex objectives and non-smooth $\ell_p$ (where $0 \leq p < 1$) regularizers. This enables our approach to seamlessly incorporate various classifiers and plausibility measures while producing sparser solutions. Our algorithm only requires differentiable data-manifold regularizers and supports box constraints for bounded feature ranges, ensuring the generated counterfactuals remain \emph{actionable}. Finally, experiments on real-world datasets demonstrate that our approach effectively produces sparse, manifold-aligned counterfactual explanations while maintaining proximity to the factual data and computational efficiency.
\end{abstract}
\section{Introduction}
\label{sec:introduction}
Machine learning models are increasingly deployed in critical decision-making scenarios, from finance and healthcare to criminal justice and hiring. While these classifiers can be highly accurate, their decision-making processes are often opaque, raising concerns about transparency, fairness, and accountability. Counterfactual explanations (CFEs) have emerged as powerful tools to provide insights into a classifier’s decision-making process by offering hypothetical \q{what-if} scenarios. Unlike explanation methods like LRP \citep{bach2015pixel} and LIME \citep{ribeiro2016should}, which identify the minimal set of features contributing to the current classification, CFEs focus on detecting the minimal set of absent features whose presence would change the classification \citep{wachter2017counterfactual}. 

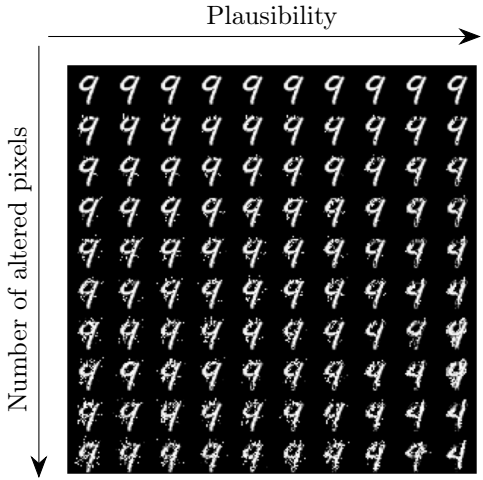
\begin{figure}
    \centering
    \begin{tikzpicture}
        \node (im) at (0, 0) {\includegraphics[width=0.35\textwidth]{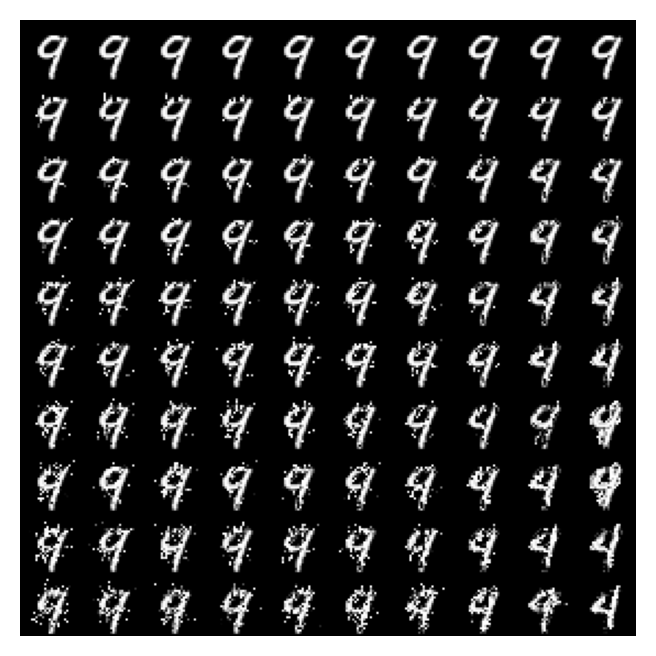}};
        \node[] (ul) at (-3.1, 3.1) {};
        \node[] (ur) at (2.9, 3.1) {};
        \node[] (ll) at (-3.1, -2.9) {};
        \draw[-{Stealth[length=3mm]}] (ul) -- (ur);
        \node[] at (0, 3.35) {Plausibility};
        \draw[-{Stealth[length=3mm]}] (ul) -- (ll);
        \node[rotate=90] at (-3.35, 0) {Number of altered pixels};
    \end{tikzpicture}
    \caption{Examples of possible CFEs for an input image of the digit 9 when changing the classification to 4: Sparsity constraints alone produce adversarial examples, while plausibility constraints lead to unrealistic CFEs. Combining both yields CFEs that are sparse yet aligned with the target class 4's data manifold.}
    \vspace{-3pt}
    \label{fig:plausibilityvssparsity}
\end{figure}

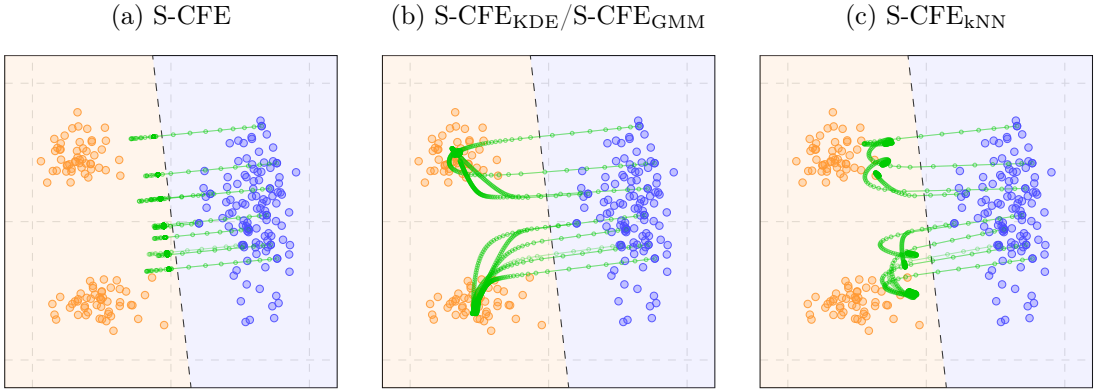
\begin{figure*}[ht]
\vspace{.3in}
\def\offs{12}
\def\imgwidth{0.6}
\centering
\def\lims{12} 
\def\mscale{0.7} 
\def\wid{6cm}
\def\hig{6cm}

\pgfplotstableread[col sep = comma]{class_1.csv}\classone
\pgfplotstableread[col sep = comma]{class_0.csv}\classzero
\pgfplotstableread[col sep = comma]{perceptron.csv}\perceptron
\pgfplotstableread[col sep = comma]{iterates_APG0.csv}\itAPG
\pgfplotstableread[col sep = comma]{iterates_APG0_KDE.csv}\itAPGKDE
\pgfplotstableread[col sep = comma]{iterates_APG0_kNN.csv}\itAPGkNN

\pgfplotsset{select coords between index/.style 2 args={
    x filter/.code={
        \ifnum\coordindex<#1\def\pgfmathresult{}\fi
        \ifnum\coordindex>#2\def\pgfmathresult{}\fi
    }
}}

\begin{tikzpicture}
\begin{axis}[
        width=\wid, height=\hig,
        grid = major,
        grid style={dashed, gray!30},
        xmin=-\lims, xmax=\lims,
        ymin=-\lims, ymax=\lims,
        ticks=none,
        legend style={at={(0.245,0.97)}, anchor=north},
        title={(a) S-CFE}
     ]

    \path[name path=below] (axis cs:-\lims,-\lims) -- (axis cs:\lims,-\lims);
    \path[name path=above] (axis cs:-\lims,\lims) -- (axis cs:\lims,\lims);
    \addplot[only marks, mark options={scale=\mscale}, blue!70!white, fill opacity=0.3, draw opacity=0.9] table[x=x, y=y] {\classone};
    \addplot[only marks, mark options={scale=\mscale}, orange!80!white, fill opacity=0.3, draw opacity=0.9] table[x=x, y=y] {\classzero};
    \addplot[name path=DB, dashed, black!80!white] table[x expr=\thisrowno{0} / 10 * \lims, y expr=\thisrowno{1} / 10 * \lims ] {\perceptron};
    \addplot[fill=orange!30!white, nearly transparent] fill between[of=DB and below];
    \addplot[fill=blue!20!white, nearly transparent] fill between[of=DB and above];

    \pgfplotstablegetcolsof{\itAPG}
    \foreach \idx in {0,...,\the\numexpr\pgfplotsretval-2\relax}{
        \addplot[mark=o, mark options={scale=0.5*\mscale}, green!80!black, fill opacity=0.3, draw opacity=0.3] table[x=x\the\numexpr\idx/2\relax, y=y\the\numexpr\idx/2\relax] {\itAPG};
        \addplot[only marks, mark options={scale=\mscale}, blue!70!white, fill opacity=0.3, draw opacity=0.9, select coords between index={0}{0}] table[x=x\the\numexpr\idx/2\relax, y=y\the\numexpr\idx/2\relax] {\itAPG};
    }
\end{axis}
\end{tikzpicture}
\hspace{10pt}
\begin{tikzpicture}
\begin{axis}[
        width=\wid, height=\hig,
        grid = major,
        grid style={dashed, gray!30},
        xmin=-\lims, xmax=\lims,
        ymin=-\lims, ymax=\lims,
        ticks=none,
        legend style={at={(0.245,0.97)}, anchor=north},
        title={(b) S-CFE$_{\textnormal{KDE}}$/S-CFE$_{\textnormal{GMM}}$}
     ]

    \path[name path=below] (axis cs:-\lims,-\lims) -- (axis cs:\lims,-\lims);
    \path[name path=above] (axis cs:-\lims,\lims) -- (axis cs:\lims,\lims);
    \addplot[only marks, mark options={scale=\mscale}, blue!70!white, fill opacity=0.3, draw opacity=0.9] table[x=x, y=y] {\classone};
    \addplot[only marks, mark options={scale=\mscale}, orange!80!white, fill opacity=0.3, draw opacity=0.9] table[x=x, y=y] {\classzero};
    \addplot[name path=DB, dashed, black!80!white] table[x expr=\thisrowno{0} / 10 * \lims, y expr=\thisrowno{1} / 10 * \lims ] {\perceptron};
    \addplot[fill=orange!30!white, nearly transparent] fill between[of=DB and below];
    \addplot[fill=blue!20!white, nearly transparent] fill between[of=DB and above];

    \pgfplotstablegetcolsof{\itAPGKDE}
    \foreach \idx in {0,...,\the\numexpr\pgfplotsretval-2\relax}{
        \addplot[mark=o, mark options={scale=0.5*\mscale}, green!80!black, fill opacity=0.3, draw opacity=0.3] table[x=x\the\numexpr\idx/2\relax, y=y\the\numexpr\idx/2\relax] {\itAPGKDE};
        \addplot[only marks, mark options={scale=\mscale}, blue!70!white, fill opacity=0.3, draw opacity=0.9, select coords between index={0}{0}] table[x=x\the\numexpr\idx/2\relax, y=y\the\numexpr\idx/2\relax] {\itAPGKDE};
    }
\end{axis}
\end{tikzpicture}
\hspace{10pt}
\begin{tikzpicture}
\begin{axis}[
        width=\wid, height=\hig,
        grid = major,
        grid style={dashed, gray!30},
        xmin=-\lims, xmax=\lims,
        ymin=-\lims, ymax=\lims,
        ticks=none,
        legend style={at={(0.245,0.97)}, anchor=north},
        title={(c) S-CFE$_{\textnormal{kNN}}$}
     ]

    \path[name path=below] (axis cs:-\lims,-\lims) -- (axis cs:\lims,-\lims);
    \path[name path=above] (axis cs:-\lims,\lims) -- (axis cs:\lims,\lims);
    \addplot[only marks, mark options={scale=\mscale}, blue!70!white, fill opacity=0.3, draw opacity=0.9] table[x=x, y=y] {\classone};
    \addplot[only marks, mark options={scale=\mscale}, orange!80!white, fill opacity=0.3, draw opacity=0.9] table[x=x, y=y] {\classzero};
    \addplot[name path=DB, dashed, black!80!white] table[x expr=\thisrowno{0} / 10 * \lims, y expr=\thisrowno{1} / 10 * \lims ] {\perceptron};
    \addplot[fill=orange!30!white, nearly transparent] fill between[of=DB and below];
    \addplot[fill=blue!20!white, nearly transparent] fill between[of=DB and above];

    \pgfplotstablegetcolsof{\itAPGkNN}
    \foreach \idx in {0,...,\the\numexpr\pgfplotsretval-2\relax}{
        \addplot[mark=o, mark options={scale=0.5*\mscale}, green!80!black, fill opacity=0.3, draw opacity=0.3] table[x=x\the\numexpr\idx/2\relax, y=y\the\numexpr\idx/2\relax] {\itAPGkNN};
        \addplot[only marks, mark options={scale=\mscale}, blue!70!white, fill opacity=0.3, draw opacity=0.9, select coords between index={0}{0}] table[x=x\the\numexpr\idx/2\relax, y=y\the\numexpr\idx/2\relax] {\itAPGkNN};
    }
\end{axis}
\end{tikzpicture}
    \vspace{-10pt}
    \vspace{.35in}
\caption{A simple dataset illustrates the need for a plausibility term in CFE algorithms. (a) Our S-CFE method without a plausibility term generates CFEs near the factual blue data points, but they remain distant from the distribution of correctly classified orange data points. (b) Our S-CFE$_{\textnormal{KDE}}$ method, which combines S-CFE with a KDE-based plausibility term, produces CFEs within high-density regions. (c) Similarly, S-CFE$_{\textnormal{kNN}}$, combining S-CFE with a $k-$NN-based plausibility term generates CFEs near the boundary of high-density regions. The green trajectory connecting the green data points represents the iterates of our S-CFE algorithm. The dashed black line represents the decision boundary of a linear classifier.}
\label{2dplausibility}
\end{figure*}

\paragraph{Basic Principles.} In essence, a CFE suggests small changes to input features \emph{(Proximity)} that could lead to a different, more favorable outcome \emph{(Validity)}. These changes must be \emph{Actionable}, meaning they should apply only to valid feature ranges and avoid unrealistic suggestions. For example, a CFE in a loan application scenario should not propose that Alice reduce her age by ten years. 
\paragraph{Plausibility.} The basic principles of CFEs —\emph{Proximity}, \emph{Validity}, and \emph{Actionability} —highlight their similarity to adversarial examples. However, a critical distinguishing conceptual feature lies in their \emph{Plausibility}. While adversarial attacks introduce small changes, known as perturbations, to mislead the classifier into incorrect predictions—often pushing the data point out of its original class distribution—CFEs aim to nudge the data point toward the target class’s distribution. This ensures that the explanations are not only effective but also grounded in plausible real-world scenarios. \cref{2dplausibility} illustrates this property using a synthetic 2D Gaussian dataset, comparing CFEs generated without a data distribution penalty to those incorporating a plausibility term, ensuring the CFEs align with the target class distribution.  
\paragraph{Sparsity.} Additionally, CFEs should modify as few features as possible to promote simplicity, a concept known as \emph{Sparsity}. Studies show that shorter explanations are easier for people to understand, making sparsity critical \citep{mothilal2020explaining, naumann2021consequence}. Combined with the proximity requirement, sparsity implies that feature changes should be minimal and low in magnitude. For instance, rather than suggesting multiple changes, a CFE might recommend that Alice raise her income by just $\$10$K to shift the loan decision from rejection to approval, as opposed to requiring a $\$50$K increase or multiple simultaneous feature adjustments. 

However, existing methods for generating CFEs often struggle to produce results that are both sparse and adhere to the data manifold, leading to unrealistic or impractical suggestions. 
To better understand the tradeoff between sparsity and plausibility, we provide an illustration in \cref{fig:plausibilityvssparsity} using an MNIST image \citep{lecun1998gradient}. The goal is to minimally alter a 9 to resemble a 4. Sparsity alone produces perturbations similar to adversarial attacks (leftmost column), while plausibility alone leads to unrealistic CFEs (rightmost column). By balancing both, we achieve simple, realistic changes, as shown in the bottom-right examples.

To capture these requirements, one would ideally solve the following optimization problem to find a CFE for a given factual data point $\boldsymbol{x}_f$
\begin{align}
\begin{split}
\min_{\boldsymbol{x} \in \text{actionable set}} &\text{counterfactual loss of $\boldsymbol{x}$}\\
+\ &\text{dist to $\boldsymbol{x}_f$}\\
+\ &\text{dist to data manifold}\\
+\ &\text{No. of feature changes}, \label{canonicalform}
\end{split}
\end{align}
which we refer to as the \emph{canonical form} of a CFE. The key technical challenge lies in the fact that the objective terms can be \emph{non-convex} and \emph{non-smooth}, due to the use of complex classifiers, intricate distance measures to the data manifold, and sparsity terms, such as those introduced by $\ell_0$ regularizers. Additionally, the actionable set imposes constraints on how much the CFE can change. Numerous methods have tackled partial aspects of this problem, often tailoring their approaches to specific classifiers, plausibility, or sparsity constraints.  For example, \cite{artelt2020convex} generate CFEs for simple linear classifiers and decision trees, while \cite{tsiourvas2024manifold} focus on ReLU neural networks (NNs) using the Local Outlier Factor (LOF) metric as a plausibility constraint. For a comprehensive survey, see \citep{verma2024counterfactual}.
\paragraph{Our Contributions:}
\begin{enumerate}
    \item We introduce S-CFE ({\bfseries S}imple {\bfseries C}ounter{\bfseries f}actual {\bfseries E}xplanations), a \emph{simple} method to solve the canonical formulation in \cref{canonicalform}, using the Accelerated Proximal Gradient (APG) method \citep{beck2009fast}.
    \item Our method is capable of handling smooth non-convex objectives and non-smooth $\ell_p$ (where $0 \leq p < 1$) regularizers. This enables our approach to seamlessly incorporate various classifiers and plausibility measures while producing \emph{sparser} solutions, that are \emph{actionable}.
    \item Extensive evaluations on real-world datasets show significant improvements over existing methods, particularly in sparsity and adherence to the data manifold, which is crucial for generating meaningful CFEs. 
\end{enumerate}
In summary, our proposed method addresses the key challenges in generating CFEs, offering a practical and effective tool for enhancing the interpretability of classifiers. This advancement is particularly valuable in safety-critical domains, where understanding and trusting classifier decisions is essential. 
    

\section{Related Work}
CFEs have seen growing interest in recent years \citep{wachter2017counterfactual, verma2024counterfactual, karimi2020survey}. Most methods enforce sparsity by optimizing weighted Manhattan or Mahalanobis distances between the factual data and the generated CFE. Only recently has the importance of plausibility in CFEs been recognized, prompting a focus on methods addressing both sparsity and plausibility. For DNN-based classifiers, \cite{dhurandhar2018explanations} were among the first to frame this as an unconstrained problem using the $\ell_1$ norm for sparsity and VAEs trained on the data distribution for plausibility. Building on \cite{van2021interpretable}, \cite{zhang2023density} replace autoencoders with a density-based distance term for plausibility, while adopting the elastic-net regularizer \citep{zou2005regularization} for sparsity. They also highlight that autoencoder-based approaches often struggle with data quality issues, undermining the robustness and credibility of CFEs. \cite{artelt2020convex} formulate a constrained optimization problem, using Gaussian Mixture Models (GMMs) to ensure high target-class density, with $\ell_1-$distance enforcing sparsity. The non-convex GMM problem is approximated by solving convex quadratic subproblems for each GMM component, leveraging simple classifiers like generalized linear classifiers and linear SVMs. The recent paper of \cite{tsiourvas2024manifold} takes a different approach, formulating a mixed-integer programming (MIP) problem for ReLU networks, using the Local Outlier Factor (LOF) for plausibility and adding a sparsity constraint. The MIP is solved efficiently by restricting the search to polytopes containing the correct class, but this framework is limited to ReLU architectures.
\section{Preliminaries}
 Assume a classification setting where $\mathcal{X} \subseteq \mathbb{R}^d$ denotes the input space, the discrete finite set $\mathcal{Y}$ denotes the set of possible class labels, and $\mathcal{D} = \{ (\boldsymbol{x}_i,y_i) \in \mathcal{X \times Y} \}_{i=1}^n$ is a dataset consisting of $n$ independent and identically distributed data points generated from a joint density $\psi: \mathcal{X \times Y} \mapsto \mathbb{R}_+$. Furthermore, we define $q(\boldsymbol{x},y):=\psi(\boldsymbol{x}|y)$, which is the corresponding density of the inputs conditioned on the given label $y$.
 
 
We let $f_{l}: \mathcal{X} \to \mathbb{R}^{|\mathcal{Y}|}$ denote a classifier that takes a $d$-dimensional sample as input and outputs logits of $|\mathcal{Y}|$ classes. The final decision is denoted by $f(\boldsymbol{x}):=\argmax_i [f_l(\boldsymbol{x})]_i$.
Furthermore, let $\theta_p : \mathcal{X} \times \mathcal{X} \mapsto \mathbb{R}_{+}$ be a distance function on $\mathcal{X}$, such as the one given by $\ell_p-$(quasi) norm $\theta_p(\boldsymbol{x}, \boldsymbol{x}^\prime):=\| \boldsymbol{x} - \boldsymbol{x}^\prime\|_p^p = \sum_{i=1}^d |x_i-x^\prime_i|^p$, for $\boldsymbol{x}, \boldsymbol{x}^\prime \in \mathbb{R}^d$ and $0<p<\infty$. For $p=0$, we define $\theta_0(\boldsymbol x,\boldsymbol x'):=\|\boldsymbol x-\boldsymbol x'\|_0$ as the cardinality of the support of $\boldsymbol x-\boldsymbol x'$. For $d \in \mathbb{N}$ let $[d] = \{ 1,...,d\}$. 
\begin{definition}[\citep{parikh2014proximal}]
    The proximal operator with respect to a (possibly non-smooth) function $g: \mathbb{R}^d \to \mathbb{R}$ is defined for any $\boldsymbol{x}^\prime \in \mathbb{R}^d$
\begin{equation}
    \begin{aligned}
        \textnormal{prox}_{\lambda g} (\boldsymbol{x}^\prime) : =\ \ & \argmin_{\boldsymbol{x}\in \mathbb{R}^d} \frac{1}{2 \lambda} \theta_2 (\boldsymbol{x},\boldsymbol{x}^\prime) + g(\boldsymbol{x}),
    \end{aligned} \nonumber
\end{equation}
where $\lambda >0$ is a given parameter.
\end{definition}
The proximal operator is particularly useful for analyzing non-smooth functions $g$ and can often be computed analytically for many such functions.
\subsection{Background on CFEs}
\begin{definition}
    \textnormal{(Closest Data-Manifold CFE)}. Given a factual data sample $\boldsymbol{x}_{f} \in \mathbb{R}^d$ such that $f(\boldsymbol{x}_{f}) = y_f,$ its closest data-manifold CFE (in terms of $\theta_2$) with respect to $f(\cdot)$ and the data manifold of the target class $y_{cf}$ is defined as a point $\boldsymbol{x}_{cf} \in \mathcal{X}$ that is the solution of the following optimization problem
    \begin{equation}
    \begin{aligned}
        \boldsymbol{x}_{cf} : =\ \ &\argmin_{\boldsymbol{x} \in \mathcal{X}}   \theta_2(\boldsymbol{x}, \boldsymbol{x}_{f}) \\
        \text{s.t.}\ \ &  \boldsymbol{x} \in \mathcal{A} \\ 
        & f(\boldsymbol{x}) = y_{cf} \\
        & q(\boldsymbol{x}, y_{cf}) \geq \tau,
    \end{aligned} \label{closestdatamanifoldcfe}
\end{equation}
where $\mathcal{A}:= \bigtimes_{i=1}^d [-\mathcal{A}_i, \mathcal{A}_i],$ for $\mathcal{A}_i \in \mathbb{R},$ denotes the value range for features, extracted from the observed dataset or given by the user, and $\tau >0$ denotes a minimum density at which we consider a sample to lie in the data manifold of the target class $y_{cf}$.
\end{definition}
The objective function in \cref{closestdatamanifoldcfe} ensures the \emph{proximity} of the generated CFE, while the three constraints account for \emph{actionability}, \emph{validity}, and \emph{plausibility}. Without the plausibility constraint, the closest CFEs often deviate significantly from the data manifold (cf. \cref{2dplausibility}), leading to unrealistic and anomalous instances. This underscores the importance of the plausibility term, which ensures that the generated CFEs remain on the data manifold.

To find the closest data-manifold CFEs while altering as few features as possible, ensuring \emph{sparsity} of the generated CFEs, the non-smooth constrain $\theta_0 (\boldsymbol{x}, \boldsymbol{x}_f) \leq m$ can be added to \cref{closestdatamanifoldcfe} as follows. 
\begin{definition}
    \textnormal{(Closest Sparse Data-Manifold CFE)}. Given a factual data sample $\boldsymbol{x}_{f} \in \mathbb{R}^d$ such that $f(\boldsymbol{x}_{f}) = y_f,$ its closest sparse data-manifold CFE with respect to $f(\cdot)$ and the data manifold of the target class $y_{cf}$ is defined as a point $\boldsymbol{x}_{cf} \in \mathcal{X}$ that is the solution of the following optimization problem
    \begin{equation}
    \begin{aligned}
        \boldsymbol{x}_{cf} : =\ \ &\argmin_{\boldsymbol{x} \in \mathcal{X}}   \theta_2(\boldsymbol{x}, \boldsymbol{x}_{f})\\
        \text{s.t.}\ \ & \boldsymbol{x} \in \mathcal{A} \\
        & f(\boldsymbol{x}) = y_{cf} \\
        & q(\boldsymbol{x}, y_{cf}) \geq \tau \\
        & \theta_0 (\boldsymbol{x}, \boldsymbol{x}_f) \leq m,
    \end{aligned} \label{closestsparsedatacfe}
\end{equation}
where $m \in \mathbb{N}$ is a parameter to explicitly control the sparsity of the generated CFE.
\end{definition}
While for the validity constraint in \cref{closestsparsedatacfe}, common measures such as checking if the CFE belongs to the target class can be used, plausibility can be assessed with a wider range of metrics. Techniques like $k$-nearest neighbors, kernel density, and Mahalanobis distance estimate whether a sample aligns with the data distribution. However, a widely used metric in the CFE literature (e.g., \citep{zhang2023density, hamman2023robust, tsiourvas2024manifold}) for assessing the similarity or anomalous nature of a generated CFE relative to the dataset $\mathcal{D} \subseteq \mathcal{X}$ is the Local Outlier Factor \citep{breunig2000lof}.
\begin{definition}[Local Outlier Factor (LOF)]
\label{LocalOutlierFactor}
    For $\boldsymbol{x} \in \mathcal{D},$ let $N_k(\boldsymbol{x})=\{ \boldsymbol{x}_1,...,\boldsymbol{x}_k \}$ be the set of $k-$Nearest Neighbors ($k-$NN) in $\mathcal{D}.$ The $k-$reachability distance $\textnormal{rd}_k$ of $\boldsymbol{x}$ with respect to $\boldsymbol{x}^\prime$ is defined by $\textnormal{rd}_k(\boldsymbol{x},\boldsymbol{x}^\prime) = \max \{\|\boldsymbol{x} - \boldsymbol{x}^\prime\|_p, d_k(\boldsymbol{x}^\prime)\},$ where $d_k(\boldsymbol{x}^\prime)$ is the $\ell_p$-norm distance between $\boldsymbol{x}^\prime$ and its $k-$th nearest instance in $\mathcal{D}.$ The $k-$local reachability of $\boldsymbol{x}$ is defined by $\textnormal{lrd}_k(\boldsymbol{x}) = |N_k(\boldsymbol{x})|(\sum_{\boldsymbol{x}_i \in N_k(\boldsymbol{x})}\textnormal{rd}_k(\boldsymbol{x},\boldsymbol{x}_i))^{-1}.$ Then, the $k-$LOF of $\boldsymbol{x}$ on $\mathcal{D}$ is defined as 
    \begin{equation}
    \begin{aligned}
        LOF_{k, \mathcal{D}}(\boldsymbol{x}) = \frac{1}{|N_k(\boldsymbol{x})|} \sum_{\boldsymbol{x}_i \in N_k(\boldsymbol{x})} \frac{\textnormal{lrd}_k (\boldsymbol{x}_i)}{\textnormal{lrd}_k(\boldsymbol{x})}.
    \end{aligned} \nonumber
\end{equation}
\end{definition}
We consider $p\in \{ 1,2, \infty \}$ and use LOF as a \emph{post-process} evaluation metric to measure whether the learned closest sparse CFE follows the data manifold. By convention, a value of $LOF_{k,\mathcal{D}}(\boldsymbol{x})$ close to $1$ indicates that $\boldsymbol{x}$ is an inlier that is aligned with the data manifold, while larger values (especially $LOF_{k,\mathcal{D}}(\boldsymbol{x}) > 1.5$) indicate that $\boldsymbol{x}$ is an outlier. 



\section{A Simple Algorithm for Generating CFEs} \label{scfeframework}
There are two main issues with solving \cref{closestsparsedatacfe} in practice. 
\begin{enumerate}
\item Setting the sparsity constraint aside for the moment, the first problem arises from the fact that the conditional distribution $q(\cdot,y)$ underlying the data $\mathcal{D}$ is often unknown. One possible solution is to incorporate plausibility constraints based on density estimates or the LOF metric. Density estimators like Gaussian mixture models (GMM) or kernel density estimation (KDE) are a common approach to estimate the density for each class based on training samples (see \cref{densityestimates}). Then, the plausibility constraint requires the resulting CFE to lie near the data manifold by enforcing $\hat{q}_{KDE} (\boldsymbol{x},y_{cf})\geq \tau$, respectively $\hat{q}_{GMM}(\boldsymbol{x},y_{cf})\geq \tau$. Similarly, the LOF metric can be used, by requiring $LOF_{k,\mathcal{D}}(\boldsymbol{x}) \leq \nu$, for a user-defined threshold $\nu$. However, this highly non-linear constraint (either using density-based estimates or the LOF metric) results in the density estimator being highly non-convex, thus further exacerbating the known computational challenges in optimizing non-linear classifiers (e.g., NNs). This is why \cite{artelt2020convex} solve the problem only for simple linear classifiers and decision trees, leveraging linearized GMMs for plausibility, resulting in convex quadratic subproblems. \cite{tsiourvas2024manifold} on the other hand, use LOF as a plausibility constraint and approximate the complex mixed-integer optimization (MIP) problem by considering it only for ReLU NNs, known for their piece-wise linear structure \citep{lee2018towards}, and solve the MIP only over polytopes consisting of points of the correct class.
\item Enforcing the sparsity constraint in \cref{closestsparsedatacfe} through the non-smooth $\theta_0-$distance is well known to be an NP-hard problem. Consequently, prior works of \cite{dhurandhar2018explanations, artelt2020convex} and \cite{zhang2023density}, mostly relax the sparsity constraint to $\theta_1$ and consider the regularized version of \cref{closestsparsedatacfe}.
\end{enumerate}
\subsection{Problem Relaxation and Solution Heuristic}
\label{solutionheuristic}
For the actionability constraint in \cref{closestsparsedatacfe}, we can utilize the indicator function such that 
\begin{equation}
    \begin{gathered}
       I_{\mathcal{A}}(\boldsymbol{x}): = \begin{cases}
         0,\ \ &\text{if}\ \boldsymbol{x} \in \mathcal{A},\\
         +\infty, \ \ &\text{otherwise.}
     \end{cases}
    \end{gathered} 
\end{equation}
To address the combinatorial optimization problem in \cref{closestsparsedatacfe} we make use of the following relaxations. First, we formulate the closest sparse data-manifold CFE problem by replacing the validity, plausibility, and sparsity constraints with penalty terms, where for sparsity instead of the $\theta_1-$distance regularization we consider any sparsity-inducing $\theta_p-$distance as a penalty. We incorporate the box constraints as indicator functions
\begin{equation}
    \begin{aligned}
        \boldsymbol{x}_{cf} : =\ \ &\argmin_{\boldsymbol{x} \in \mathbb{R}^d}   \theta_2(\boldsymbol{x}, \boldsymbol{x}_{f}) + I_{\mathcal{A}}(\boldsymbol{x}) +\gamma \mathcal L_f(\boldsymbol{x}, y_{cf}) \\
        & \ \ \ \  - \tau \hat{q}(\boldsymbol{x},y_{cf}) + \beta \theta_{p}(\boldsymbol x , \boldsymbol{x}_{f}), 
    \end{aligned} \label{lagrangianclosestsparsedatacfe}
\end{equation}
where we consider an estimate $\hat{q}(\cdot,y_{cf})$ for the density of target class $y_{cf}$ in $\mathcal{X}$, and $\mathcal L_f$ is a suitable (differentiable) classification loss. $\gamma, \tau > 0$ denote tradeoff parameters for validity and plausibility. The only requirement for the plausibility term $\hat{q}(\cdot, y_{cf})$ is to be differentiable so that we are able to learn from its gradient information. 

We solve the problem in \cref{lagrangianclosestsparsedatacfe} by making use of a commonly used algorithm for non-convex and non-smooth programs, based on accelerated proximal gradient (APG) method \citep{beck2009fast}.

We start by denoting $h(\boldsymbol{x},y_{cf}) :=  \theta_2(\boldsymbol{x}, \boldsymbol{x}_{f}) +\gamma \mathcal L_f(\boldsymbol{x}, y_{cf}) - \tau \hat{q}(\boldsymbol{x},y_{cf})$ and $g_{p}(\boldsymbol{x}): =I_{\mathcal{A}}(\boldsymbol{x}) + \beta \theta_{p}(\boldsymbol x , \boldsymbol{x}_{f})$. Assuming $h(\cdot, y_{cf})$ is a smooth, possibly non-convex function, whose gradient has Lipschitz constant $L$, we make a quadratic approximation $\tilde{h}_{L}(\boldsymbol{x},y_{cf})$ to $h(\boldsymbol{x},y_{cf})$ and replace $\nabla^{2}h(\boldsymbol{x},y_{cf})$ by $\frac{L}{2} I$.
Given iterate $\boldsymbol{x}^t$ of the algorithm, it holds that
\begin{equation}
    \begin{aligned}
        \boldsymbol{x}^{t+1} := \ \ &\argmin_{\boldsymbol{x} \in \mathbb{R}^d}    \tilde{h}_L(\boldsymbol{x}^t,y_{cf}) + g_{p}(\boldsymbol{x}) \\
        = \ \ &  \argmin_{\boldsymbol{x} \in \mathbb{R}^d} \nabla_{\boldsymbol{x}^t} h (\boldsymbol{x}^t,y_{cf})^\top (\boldsymbol{x}-\boldsymbol{x}^t) + \frac{L}{2} \theta_2(\boldsymbol{x}, \boldsymbol{x}^t) \\ &  \ \ \ \ + g_{p}(\boldsymbol{x})\\
        = \ \ &  \argmin_{\boldsymbol{x} \in \mathbb{R}^d} \frac{L}{2} \theta_2\left(\boldsymbol{x}, \boldsymbol{x}^t - \frac{1}{L} \nabla_{\boldsymbol{x}^t} h (\boldsymbol{x}^t,y_{cf})\right) \\ & \ \ \ \ + g_{p}(\boldsymbol{x})\\
        = \ \ & \textnormal{prox}_{\frac{1}{L}  g_{p} } \left(\boldsymbol{x}^t - \frac{1}{L} \nabla_{\boldsymbol{x}^t} h (\boldsymbol{x}^t,y_{cf})\right).
    \end{aligned} \label{quadraticapproxprox}
\end{equation}
In practice, the inverse Lipschitz constant is further replaced by a step size sequence $(\sigma_t)_{t \in \mathbb{N}}$ \citep{karimi2016linear}. In order to solve the proximal operator in \cref{quadraticapproxprox} for the function $g_{p}(\cdot)$, we first need to compute $\nabla_{\boldsymbol{x}^t} h(\boldsymbol{x}^t,y_{cf})$, which we will denote as $\nabla_{\boldsymbol{x}} h(\boldsymbol{x},y_{cf})$ for simplicity in the following sections.
\subsubsection{Computing \texorpdfstring{$\nabla_{\boldsymbol{x}}h(\boldsymbol{x},y_{cf})$}{Lg}}
\label{gradientofh}
As long as we use differentiable terms for proximity, classifier loss function, and the density term, any suitable differentiation technique can be applied to compute $\nabla_{\boldsymbol{x}}h(\boldsymbol{x},y_{cf})$, and we use backpropagation in our experiments. 

As for $\hat{q}(\boldsymbol{x},y_{cf})$, we adopt traditional differentiable estimates, such as KDE ($\hat{q}_{KDE} (\boldsymbol{x},y_{cf})$) and GMM ($\hat{q}_{GMM} (\boldsymbol{x},y_{cf})$). 
For completeness, we also consider a recently proposed density term based on the LOF metric. \cite{zhang2023density} consider an approximation to LOF given by density gravity on an instance - $G(\boldsymbol{x})$ (see Definition \ref{densitygravity}).
A suitable plausibility term then minimizes the $\ell_2-$distance between the CFE $\boldsymbol{x}$ and the density gravity point $G_{y_{cf}} (\boldsymbol{x}_f)$, generated by a convex combination of $k-$nearest neighbors of the factual data point belonging to the target class $y_{cf}$, weighted by their local density 
\begin{equation}
    \begin{aligned}
        \hat{q}_{kNN} (\boldsymbol{x},y_{cf}):= -\left\|\boldsymbol{x} - G_{y_{cf}} (\boldsymbol{x}_f)\right\|_2.
    \end{aligned} 
    \nonumber
\end{equation}
The choice of plausibility term ($\hat{q}_{KDE},\hat{q}_{GMM}$, or $\hat{q}_{kNN}$) leads to different algorithm variants: S-CFE$_{\textnormal{KDE}}$, S-CFE$_{\textnormal{GMM}}$, and S-CFE$_{\textnormal{kNN}}$.

\subsection{Computing the Proximal Operator} \label{s-cfe}
After computing $\nabla_{\boldsymbol{x}}h(\boldsymbol{x},y_{cf})$ as outlined in \cref{gradientofh}, we define $S_{\sigma} (\boldsymbol{x},y_{cf}) = \boldsymbol{x} - \sigma \nabla_{\boldsymbol{x}} h (\boldsymbol{x},y_{cf})$. The solution to the proximal operator in \cref{quadraticapproxprox} for $p=0$ is detailed in \cite{zhu2021sparse}. Notably, our approach is quite general; similar results can be derived for any $\theta_p$, with a straightforward proximal operator (e.g., $p \in \{1/2, 2/3\}$ \citep{cao2013fast, 2023_SadikuWagnerPokutta_Groupwisesparseattacks}). The method proposed by \cite{zhang2023density} is a specific case using $\theta_1$ and the density gravity term for plausibility.

We summarize our procedure in \cref{SCFE}. This method generates sparse, manifold-aligned CFEs, but we cannot precisely control the number of features that change.
\subsubsection{Constraining the Sparsity}
\label{sparsityconstrained}
Rather than penalizing sparsity in \cref{lagrangianclosestsparsedatacfe}, we can regularize using the indicator function of the sparsity constraint for improved control. This approach begins with the observation that
\begin{equation}
    \begin{gathered}
    I_{\theta_{p}(\boldsymbol x , \boldsymbol{x}_{f})\leq m}(\boldsymbol x):=\begin{cases}
        0,\quad &\text{if}\ \theta_{p}(\boldsymbol x , \boldsymbol{x}_{f})\leq m,\\
        + \infty,\quad &\text{otherwise}.
    \end{cases}
    \nonumber
\end{gathered} 
\end{equation}
Thus, problem in \cref{lagrangianclosestsparsedatacfe} can be reframed as 
\begin{equation}
    \begin{aligned}
        \boldsymbol{x}_{cf} : =\ \ &\argmin_{\boldsymbol{x} \in \mathbb{R}^d}   \theta_2(\boldsymbol{x}, \boldsymbol{x}_{f}) + I_{\mathcal{A}}(\boldsymbol{x}) +\gamma \mathcal L_f(\boldsymbol{x}, y_{cf}) \\
        & \ \ \ \  - \tau \hat{q}(\boldsymbol{x},y_{cf}) + \beta I_{\theta_{p}(\boldsymbol x , \boldsymbol{x}_{f})\leq m}(\boldsymbol x).
    \end{aligned} \label{constrainedclosestsparsedatacfe}
\end{equation}
Since the new function $g_{p}(\boldsymbol{x}):= I_{\mathcal{A}}(\boldsymbol{x})+ \beta I_{\theta_{p}(\boldsymbol x , \boldsymbol{x}_{f})\leq m}(\boldsymbol x)$ is an indicator function, its proximal operator for $p=0$ coincides with the projection onto the intersection $\{\theta_0(\boldsymbol{x}, \boldsymbol{x}_f) \leq m\} \cap \mathcal{A}$ of a $0-$norm ball and our box constraints. A closed-form solution for this projection was derived by \cite{croce2019sparse} and is given by
\begin{align*}
    \left[ P_{\{\theta_0(\boldsymbol{x}, \boldsymbol{x}_f) \leq m\} \cap \mathcal{A}}(S_{\sigma}(\boldsymbol{x},y_{cf})) \right]_i&=\begin{cases}
        z_i,\quad &\text{if}\ i\in Q,\\
        0,\quad &\text{otherwise},
    \end{cases}\\
    \boldsymbol z&=\Pi_{\mathcal{A}}(S_{\sigma}(\boldsymbol{x},y_{cf})),\\
    Q&=\text{argtopk}(\boldsymbol v, m),
\end{align*}
where $\boldsymbol v=\boldsymbol w\odot \boldsymbol w-(\boldsymbol w-\boldsymbol z)\odot(\boldsymbol w-\boldsymbol z)$ with $\boldsymbol w=\boldsymbol x-\boldsymbol x_f$ and $\odot$ the element-wise product, argtopk$(\boldsymbol v,m)$ represents the indices corresponding to the $m$ largest absolute values of the entries of $\boldsymbol v$, and $\Pi_{\mathcal{A}} (\boldsymbol{x})= \argmin_{\boldsymbol{x}} \{\|\boldsymbol{x}^\prime-\boldsymbol{x}\|_2 \ \big| \ \boldsymbol{x}^\prime \in \mathcal{A} \}$.
\begin{remark}  Since $g_0(\boldsymbol{x}):= I_{\mathcal{A}}(\boldsymbol{x})+ \beta I_{\theta_{0}(\boldsymbol x , \boldsymbol{x}_{f})\leq m}(\boldsymbol x)$ is a proper and lower semicontinuous function, the convergence of APG to a critical point of the minimization problem specified in  \cref{constrainedclosestsparsedatacfe} can be assured (even for non-convex and non-smooth $g_0(\cdot)$), under some mild conditions \citep{li2015accelerated}.
\end{remark}
\begin{algorithm}[t]
\small
\caption{S-CFE: Simple Counterfactual Explanations}
\begin{algorithmic}[1]
\REQUIRE Classifier $f$, density penalty $\hat{q}$, sparsity penalty $\theta_p$, classification loss function $\mathcal L_f$, feature ranges $\mathcal{A}$, original point $\boldsymbol{x}_f\in\mathbb R^d$, target label $y_{cf}$, parameters of the objective function $\beta, \gamma, \tau$, extrapolation parameters $(\alpha_t)$, step sizes $(\sigma_t)$, number of iterations $T$.
\setstretch{1.3}
\ENSURE $\boldsymbol{x}_{cf}\in\mathbb R^d$ the CFE of $\boldsymbol{x}_f$.\\

\STATE Initialize $\boldsymbol x^0, \boldsymbol z^0 \leftarrow \boldsymbol 0\in\mathbb R^d$.
\STATE $h(\boldsymbol x, y_{cf}):=\theta_2(\boldsymbol x,\boldsymbol x_f)+\gamma\mathcal L_f(\boldsymbol x, y_{cf})-\tau\hat{q}(\boldsymbol x,y_{cf})$
\STATE $g_{p}(\boldsymbol{x}): =I_{\mathcal{A}}(\boldsymbol{x}) + \beta \theta_{p}(\boldsymbol x , \boldsymbol{x}_{f})$
\FOR{$t=0,...,T-1$}
    \STATE $\boldsymbol r^{t+1}\leftarrow\nabla_{\boldsymbol z^{t}}h(\boldsymbol z^{t}, y_{cf})$
    \STATE $S_{\sigma_{t+1}}(\boldsymbol z^{t+1}, y_{cf})\leftarrow \boldsymbol z^{t}-\sigma_{t+1} \boldsymbol r^{t+1}$
    \STATE $\boldsymbol x^{t+1}\leftarrow\text{prox}_{\sigma_{t+1} g_p}(S_{\sigma_{t+1}}(\boldsymbol z^{t+1}, y_{cf}))$
    \STATE $\boldsymbol z^{t+1}\leftarrow \boldsymbol x^{t+1} + \alpha_{t+1}(\boldsymbol x^{t+1}-\boldsymbol x^{t})$
\ENDFOR
\STATE \textbf{return} $\boldsymbol x^T=: \boldsymbol{x}_{cf}$
\end{algorithmic}
\label{SCFE}
\end{algorithm}

\section{Experiments}
We conduct experiments on four real-world datasets to validate our methods against various benchmarks. All computations are performed using Python 3.12 \citep{van2009introduction}, PyTorch 2.4.1 \citep{paszke2019pytorch}, and Scikit-learn 1.5.2 \citep{buitinck2013api}, utilizing eight cores of an Intel Xeon Gold 6342 CPU @ 2.80GHz with 16GB of memory. Our codes are available at \href{https://github.com/wagnermoritz/SCFE}{https://github.com/wagnermoritz/SCFE}.

\subsection{Setup}
\label{setup}
\subsubsection{Datasets}
\label{datasets}
We use the Boston Housing dataset ($d=12$) \citep{boston1978housing} to predict whether the median house value exceeds a threshold, the Breast Cancer Wisconsin dataset ($d=30$) \citep{william1995breast} to classify malignant versus benign tumors, Wine dataset \citep{aeberhard1992wine} for wine quality classification, and the MNIST dataset \citep{asuncion2007uci} for handwritten digit classification. All continuous features are scaled using a min-max scaler to the $[0,1]$ range. MNIST consists of 60,000 training images and 10,000 test images, while the other datasets are split with 100 test points.
\subsubsection{Method Implementation} \label{methods}
For the experiments, we consider a $2-$layer, densely connected ReLU network with 20 neurons per hidden layer for the smaller datasets, and a convolutional neural network (CNN) consisting of two convolutional layers followed by three densely connected layers for the MNIST dataset, as the underlying machine learning models that give predictions. We train each network with a learning rate of $10^{-3}$ using the Adam optimizer. The densely connected networks are trained using a batch size of 32 for 20 epochs and the CNN is trained using a batch size of 128 for 80 epochs. All classifiers reach an accuracy above 95\% on the test data.

Following \emph{Alibi Explain} \citep{klaise2021alibi}, we conduct a logarithmic grid search over $[10^{-3}, 10^3]^2$ for $\beta$ and $\tau$, and perform a section search with 10 steps for $\gamma$ for each test instance. For binary classifiers $f:\mathcal X\rightarrow[0,1]$ we utilize the classification loss with $y_{cf} \in \{0, 1\}$
\begin{align*}
    \mathcal L_f(\boldsymbol x,y_{cf}):=\max\{(1-2y_{cf})f(\boldsymbol x), -c\},
\end{align*}
and for multi-class classifiers $f:\mathcal X\rightarrow\mathbb R^{|\mathcal{Y}|}$ with $y_{cf}\in\{1,...,|\mathcal{Y}|\}$ we use
\begin{align*}
    \mathcal L_f(\boldsymbol x, y_{cf}):=\max\{\max\{f_i(\boldsymbol x)\ |\ i\neq y_{cf}\}-f_{y_{cf}}(\boldsymbol x),-c\},
\end{align*}
where $\boldsymbol x$ a datapoint of class $y_{cf}$ and $c$ is the cut-off parameter for the hinge loss. Similar to most adversarial attacks utilizing the Hinge Loss (or Carlini-Wagner loss) \citep{carlini2017towards}, we set its threshold to $c = 0$.
Following \cite{zhang2023density}, in the S-CFE$_{\textnormal{kNN}}$ approach, we use 3, 4, and 5 for the grid search of nearest neighbors $k$, as higher values tend to confuse the classifier and degrade performance (see \cref{manyneighborsissue}).

In line with \cite{beck2009fast}, we use the sequence $(\alpha_t)$ of extrapolation parameters defined by
\begin{align}
    \beta_t=\frac{1}{2}\left(1+\sqrt{1+4 \beta^2_{t-1}}\right),\ \ \alpha_t=\frac{\beta_t-1}{\beta_{t+1}},
\end{align}
with $\beta_0=0$. The sequence of step sizes $(\sigma_t)$ is given by an initial step size $\sigma_0$ and a square-root decay
\begin{align*}
    \sigma_{t+1}=\sigma_t\sqrt{1-\frac{t}{t_{\max}}},
\end{align*}
where $t_{\max}$ is the maximum number of iterations. We run all the methods for a total of $200$ iterations.

\subsubsection{Evaluation Metrics}
\label{evaluation}
In our experiments, we report validity, defined as the ratio of CFEs with the desired class label multiplied by $100\%$. We measure proximity using the $\ell_2-$norm, $\|\boldsymbol{x}_{cf} - \boldsymbol{x}_{f}\|_2$, and sparsity using the $\ell_0-$(quasi) norm, $\|\boldsymbol{x}_{cf} - \boldsymbol{x}_{f}\|_0$, where $\boldsymbol{x}_{cf}$ is the generated CFE and $\boldsymbol{x}_{f}$ the original factual point.
To assess plausibility, we use the LOF metric (Definition \ref{LocalOutlierFactor}) to determine how much a generated CFE deviates from the data manifold, based on its $k$ nearest neighbors in $\mathcal{D}$. The scikit-learn implementation of LOF is used with default parameters.  Additionally, we report the average runtime per method.

\subsection{Results} \label{Results}

\begin{table*}[t]
    \caption{CFEs for linear classifiers on the Boston Housing and Wine datasets. Dimensionality was reduced to 8 using PCA, with 100 test points assessed. Compute time is reported in seconds per 100 CFEs. The best values for each dataset are highlighted. The method proposed by \cite{artelt2020convex} is referred to as PCFE.}
    \centering
    \small
    \begin{tabular}{ccccccc}
    \toprule
        Dataset & Method & Validity (std) & $\ell_2$ (std) & $\ell_0$ (std) & LOF (std) & Time\\
    \midrule
        \multirow{4}{*}{\begin{tabular}{c}Housing\\8 features\end{tabular}} & S-CFE$_{KDE}$ & \textbf{100} (0.00) & 3.06 (1.42) & \textbf{1.00} (0.00) & 1.22 (0.27) & 11.8 \\
        & S-CFE$_{GMM}$ & \textbf{100} (0.00) & 2.62 (1.25) & \textbf{1.00} (0.00) & \textbf{1.20} (0.27) & 12.1 \\
        & S-CFE$_{k-\text{NN}}$ & \textbf{100} (0.00) & 2.51 (1.19) & \textbf{1.00} (0.00) & 1.34 (0.58) & \textbf{4.61} \\
        & PCFE & \textbf{100} (0.00) & \textbf{1.61} (1.03) & 1.13 (0.87) & 1.31 (0.57) & 26.7 \\
    \midrule
        \multirow{4}{*}{\begin{tabular}{c}Wine\\8 features\end{tabular}} & S-CFE$_{KDE}$ & \textbf{100} (0.00) & 2.10 (1.10) & \textbf{1.00} (0.00) & \textbf{0.98} (0.01) & 9.88 \\
        & S-CFE$_{GMM}$ & \textbf{100} (0.00) & 2.09 (1.10) & \textbf{1.00} (0.00) & \textbf{0.98} (0.02) & 11.4 \\
        & S-CFE$_{k-\text{NN}}$ & \textbf{100} (0.00) & 2.15 (1.12) & \textbf{1.00} (0.00) & 0.99 (0.02) & \textbf{4.50} \\
        & PCFE & \textbf{100} (0.00) & \textbf{1.39} (1.02) & 1.37 (0.96) & 0.99 (0.02) & 22.3 \\
    \bottomrule
    \end{tabular}
    \label{tab:linear}
\end{table*}

\begin{table*}[t]
    \caption{CFEs for DNN classifiers on the Boston Housing and Wine datasets, and for a CNN classifier on the MNIST dataset. Evaluated on 1000 test points for MNIST and 100 test points for the other two datasets. The compute time is given in seconds per 100 CFEs. The best values for each dataset are highlighted. The methods proposed by \cite{zhang2023density} and \cite{dhurandhar2018explanations} are referred to as DCFE and CEM, respectively.}
    \centering
    \small
    \begin{tabular}{ccccccc}
    \toprule
       Dataset & Method & Validity (std) & $\ell_2$ (std) & $\ell_0$ (std) & LOF (std) & Time\\
    \midrule
        \multirow{5}{*}{\begin{tabular}{c}Housing\\12 features\end{tabular}} & S-CFE$_{KDE}$ & \textbf{100} (0.00) & \textbf{2.59} (1.21) & \textbf{2.00} (0.00) & 1.23 (0.29) & 12.7 \\
        & S-CFE$_{GMM}$ & \textbf{100} (0.00) & 2.91 (1.38) & \textbf{2.00} (0.00) & \textbf{1.12} (0.26) & 13.3 \\
        & S-CFE$_{k_\text{NN}}$ & \textbf{100} (0.00) & 3.64 (1.73) & \textbf{2.00} (0.00) & 1.17 (0.31) & 5.85 \\
        & DCFE & \textbf{100} (0.00) & 3.50 (1.68) & 6.86 (1.42) & 1.27 (0.38) & \textbf{5.33} \\
        & CEM & 94.0 (0.23) & 2.93 (2.23) & 2.99 (1.17) & 1.36 (0.60) & 7.51\\
    \midrule
        \multirow{5}{*}{\begin{tabular}{c}Wine\\13 features\end{tabular}} & S-CFE$_{KDE}$ & \textbf{100} (0.00) & 3.31 (1.16) & \textbf{2.00} (0.00) & 0.99 (0.01) & 12.4 \\
        & S-CFE$_{GMM}$ & \textbf{100} (0.00) & 3.44 (1.09) & \textbf{2.00} (0.00) & \textbf{0.98} (0.02) & 13.1 \\
        & S-CFE$_{k-\text{NN}}$ & \textbf{100} (0.00) & 4.04 (1.59) & \textbf{2.00} (0.00) & 1.01 (0.07) & 5.80 \\
        & DCFE & \textbf{100} (0.00) & \textbf{3.21} (2.70) & 7.13 (1.31) & 1.03 (0.18) & \textbf{4.95} \\
        & CEM & 92.0 (0.29) & 5.40 (3.25) & 5.14 (2.68) & 1.07 (0.14) & 5.71 \\
    \midrule
        \multirow{3}{*}{\begin{tabular}{c}MNIST\\784 features\end{tabular}} & S-CFE$_{GMM}$ & 99.1 (0.09) & \textbf{6.74} (2.92) & \textbf{25.0} (0.00) & \textbf{1.21} (0.18) & 55.3 \\
        & S-CFE$_{k-\text{NN}}$ & \textbf{99.8} (0.04) & 7.04 (2.99) & \textbf{25.0} (0.00) & 1.30 (0.22) & 13.1 \\
        & DCFE & 99.3 (0.08) & 8.06 (3.48) & 118 (6.30) & 1.32 (2.24) & \textbf{11.8} \\
    \bottomrule
    \end{tabular}
    \label{tab:nonlinear}
\end{table*}

\begin{figure*}
\centering 
    \footnotesize
    \definecolor{plotcolor1}{HTML}{1b9e77}
\definecolor{plotcolor2}{HTML}{d95f02}
\definecolor{plotcolor3}{HTML}{7570b3}
\definecolor{plotcolor4}{HTML}{e7298a}

\begin{tikzpicture}
\pgftransformscale{1}
\begin{groupplot}[group style={group size= 4 by 1, horizontal sep=0.8cm, vertical sep=1.3cm},height=4.5cm,width=5cm, , legend columns=-1]
    \nextgroupplot[
    x label style={at={(axis description cs:0.5,0.0)},anchor=north},
    xlabel = {$\ell_2-$norm input},
    y label style={at={(axis description cs:0.25,0.5)},anchor=south},
    ylabel = {$\ell_2-$norm output},
    xmajorgrids=true,
    ymajorgrids=true,
    every x tick scale label/.style={at={(xticklabel cs:0.5)}, color=white},
    title={Boston Housing, $p=0$},
    legend style={{at=(0.025,0.025)}, anchor=south west},
    legend to name=legend_plt,
    cycle multiindex list={plotcolor1, plotcolor2, plotcolor3, plotcolor4, plotcolor5, plotcolor6, plotcolor7\nextlist mark list}]
    
    \addplot
        table[x=l2,y=l2_no_plaus,col sep=comma]{BostonRobustness0.csv};
    \addlegendentry{No Plaus.};
    \addplot
        table[x=l2,y=l2_kde,col sep=comma]{BostonRobustness0.csv};
    \addlegendentry{KDE};
    \addplot
        table[x=l2,y=l2_gmm,col sep=comma]{BostonRobustness0.csv};
    \addlegendentry{GMM};
    \addplot
        table[x=l2,y=l2_knn,col sep=comma]{BostonRobustness0.csv};
    \addlegendentry{$k$-NN};
    \coordinate (top) at (rel axis cs:0,1);
    
    \nextgroupplot[
    x label style={at={(axis description cs:0.5,0.0)},anchor=north},
    xlabel = {$\ell_2-$norm input},
    xmajorgrids=true,
    ymajorgrids=true,
    every x tick scale label/.style={at={(xticklabel cs:0.5)}, color=white},
    title={Boston Housing, $p=1$},
    legend style={{at=(0.025,0.025)}, anchor=south west},
    legend to name=legend_plt,
    cycle multiindex list={plotcolor1, plotcolor2, plotcolor3, plotcolor4, plotcolor5, plotcolor6, plotcolor7\nextlist mark list}]
    
    \addplot
        table[x=l2,y=l2_no_plaus,col sep=comma]{BostonRobustness1.csv};
    \addlegendentry{No Plaus.};
    \addplot
        table[x=l2,y=l2_kde,col sep=comma]{BostonRobustness1.csv};
    \addlegendentry{KDE};
    \addplot
        table[x=l2,y=l2_gmm,col sep=comma]{BostonRobustness1.csv};
    \addlegendentry{GMM};
    \addplot
        table[x=l2,y=l2_knn,col sep=comma]{BostonRobustness1.csv};
    \addlegendentry{$k$-NN};

    \nextgroupplot[
    x label style={at={(axis description cs:0.5,0.0)},anchor=north},
    xlabel = {$\ell_2-$norm input},
    xmajorgrids=true,
    ymajorgrids=true,
    every x tick scale label/.style={at={(xticklabel cs:0.5)}, color=white},
    title={Wisconsin, $p=0$},
    legend style={{at=(0.025,0.025)}, anchor=south west},
    legend to name=legend_plt,
    cycle multiindex list={plotcolor1, plotcolor2, plotcolor3, plotcolor4, plotcolor5, plotcolor6, plotcolor7\nextlist mark list}]
    
    \addplot
        table[x=l2,y=l2_no_plaus,col sep=comma]{WisconsinRobustness0.csv};
    \addlegendentry{No Plaus.};
    \addplot
        table[x=l2,y=l2_kde,col sep=comma]{WisconsinRobustness0.csv};
    \addlegendentry{KDE};
    \addplot
        table[x=l2,y=l2_gmm,col sep=comma]{WisconsinRobustness0.csv};
    \addlegendentry{GMM};
    \addplot
        table[x=l2,y=l2_knn,col sep=comma]{WisconsinRobustness0.csv};
    \addlegendentry{$k$-NN};

    \nextgroupplot[
    x label style={at={(axis description cs:0.5,0.0)},anchor=north},
    xlabel = {$\ell_2-$norm input},
    xmajorgrids=true,
    ymajorgrids=true,
    every x tick scale label/.style={at={(xticklabel cs:0.5)}, color=white},
    title={Wisconsin, $p=1$},
    legend style={{at=(0.025,0.025)}, anchor=south west},
    legend to name=legend_plt,
    cycle multiindex list={plotcolor1, plotcolor2, plotcolor3, plotcolor4, plotcolor5, plotcolor6, plotcolor7\nextlist mark list}]
    
    \addplot
        table[x=l2,y=l2_no_plaus,col sep=comma]{WisconsinRobustness1.csv};
    \addlegendentry{No Plaus.};
    \addplot
        table[x=l2,y=l2_kde,col sep=comma]{WisconsinRobustness1.csv};
    \addlegendentry{KDE};
    \addplot
        table[x=l2,y=l2_gmm,col sep=comma]{WisconsinRobustness1.csv};
    \addlegendentry{GMM};
    \addplot
        table[x=l2,y=l2_knn,col sep=comma]{WisconsinRobustness1.csv};
    \addlegendentry{$k$-NN};
    
    \coordinate (bot) at (rel axis cs:0.815,0);
\end{groupplot}
\path (top)--(bot) coordinate[midway] (group center);
\node[inner sep=0pt] at ([yshift=-0.5cm, xshift=0.3cm] group center |- current bounding box.south) {\pgfplotslegendfromname{legend_plt}};
\normalsize
\end{tikzpicture}
    \vspace{15pt}
\caption{Robustness of the different methods. The distance of the input data points to the original data points on the $x$-axis and the distance of the generated CFEs to the CFE generated from the original data points on the $y$-axis. Tested on 100 data points from each data set.}
\label{robustnessplotsmain}
\vspace{-5pt}
\end{figure*}
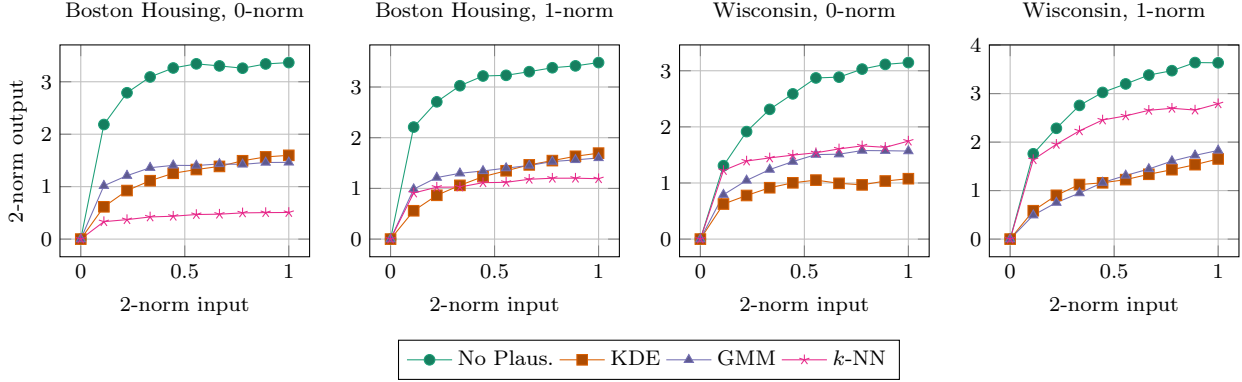

Proximity, sparsity, and plausibility are often conflicting objectives, as data near the decision boundary is typically sparse, and small shifts across it can yield implausible CFEs \citep{van2021interpretable, dandl2024countarfactuals}. By applying our S-CFE algorithm with various plausibility terms, and the constrained $\theta_0-$metric via indicator functions to achieve sparsity, we are able to generate sparse, plausible CFEs that remain close to the factual data points and maintain the computation speed. 

\subsubsection{Quantitative Evaluation} \label{quanteval}
As \cite{artelt2020convex} consider only linear classifiers and decision trees, \cref{tab:linear} reports results for logistic regression classifiers on the Housing and Wine datasets, including the CFE validity, average proximity, average sparsity, plausibility (measured via LOF), and generation time for both CFE algorithms. \cref{tab:nonlinear} presents the same metrics for the Housing, Wine, and MNIST datasets using a CNN classifier, comparing variants of our method against \citep{dhurandhar2018explanations, zhang2023density}.

In terms of validity, S-CFE consistently outperforms other methods (e.g., on MNIST), generating the desired CFE for nearly all original points. Thanks to our formulation in \cref{constrainedclosestsparsedatacfe}, we control the number of altered features, setting it to 1 in \cref{tab:linear} and 2 in \cref{tab:nonlinear}, producing the sparsest CFEs. The integration of differentiable plausibility terms guides the search towards high-density regions, resulting in the lowest LOF. This allows us to generate superior sparse, manifold-aligned CFEs compared to benchmark works of \cite{dhurandhar2018explanations, artelt2020convex} and \cite{zhang2023density}. Additionally, we achieve comparable or slightly better proximity to factual data while maintaining low computation time, underscoring the simplicity of our method.
See \cref{tab:linearextended} and \cref{tab:nonlinearextended} in \cref{additionalexperiments} for additional experiments.

Note that CEM \citep{dhurandhar2018explanations} can only be applied to binary classification datasets, where the goal is to find CFEs for a certain target class. Their pertinent positives represent the minimal set of features required to maintain the current classification, while pertinent negatives are the minimal set needed to change the classification. In datasets like MNIST, the altered class could be any of the nine remaining classes (e.g., a \q{9} could change into any class other than \q{4}), making CEM unsuitable for targeted CFEs in multi-class scenarios.

\cite{tsiourvas2024manifold}, on the other hand, reference a link to an empty repository, hindering result reproducibility. The MIP formulation of \cite{tsiourvas2024manifold} faces scalability issues with complex networks, addressed by limiting the search to a few live polytopes from the correct class. However, this partitioning is restricted to ReLU NNs and often compromises plausibility.  \cref{miplivevsours} conceptually shows our method outperforming MIP-Live in terms of plausibility on their simple example, and we conjecture this advantage extends to higher dimensions, as their approach overly constrains the search space, leading to a poor sparsity-plausibility tradeoff. Moreover, limiting the method to ReLU networks ignores the growing success of other architectures with other nonlinearities than ReLU (e.g., Swin transformers in Computer Vision using GELU activations \citep{liu2021swin}). In contrast, our approach supports any architecture that utilizes standard gradient computation techniques like backpropagation. Consequently, we exclude MIP-Live from our experiments.

\subsubsection{Robustness of Plausible CFEs to Input Manipulations} \label{robustness}
CFEs without plausibility constraints have been shown to diverge significantly with even minor input perturbations, underscoring their lack of robustness \citep{slack2021counterfactual}. This presents a challenge for CFEs, as two similar individuals may receive drastically different explanations. In contrast, incorporating plausibility constraints improves robustness against such input shifts, enhancing the individual fairness of CFEs \citep{artelt2021evaluating, zhang2023density}. \cref{robustnessplotsmain} demonstrates that sparse, manifold-aligned CFEs generated with various plausibility regularizers further enhance robustness against input shifts. For additional results across various datasets and different sparsity modes, refer to \cref{robustnessplots} in \cref{additionalexperiments}.

\section{Conclusion and Discussion} \label{Discussion}
We introduced \q{S-CFE}, a novel yet simple framework for generating sparse and plausible counterfactual explanations. Our method is based on proximal gradient techniques for non-convex and non-smooth optimization, offering enhanced control over feature changes and leveraging density estimates to ensure plausibility. Extensive experiments demonstrate that S-CFE outperforms existing methods in producing sparse, plausible CFEs while maintaining proximity to input data and computational efficiency. 
\paragraph{Limitations and Broader Impact:}
Sparse plausible CFEs highlight which feature changes may lead to different predictions but offer no guidance on real-world interventions for achieving the desired outcome, which requires causal knowledge. \emph{Improving} the underlying target is more important than merely gaining predictor acceptance \citep{tsirtsis2020decisions}. For example, altering symptoms may change a COVID-19 diagnosis, but not the actual infection status \citep{konig2023improvement}. Our method acts as an adversary, guiding users toward simple changes for predictor acceptance without improving the real-world state. Future work will explore training S-CFE directly on data rather than relying on classifier predictions.

\section*{Acknowledgements}
This research was partially supported by the Deutsche Forschungsgemeinschaft (DFG, German Research Foundation) under Germany’s Excellence Strategy through the Berlin Mathematics Research Center MATH+ (EXC-2046/1, Project ID: 390685689). Additional support was provided by the Research Campus MODAL, funded by the German Federal Ministry of Education and Research (BMBF, Grant No. 05M14ZAM).

\bibliographystyle{plainnat}
\bibliography{literature}

\section*{Checklist}


 \begin{enumerate}

 \item For all classifiers and algorithms presented, check if you include:
 \begin{enumerate}
   \item A clear description of the mathematical setting, assumptions, algorithm, and/or classifier. \textbf{Yes}
   \item An analysis of the properties and complexity (time, space, sample size) of any algorithm. \textbf{Yes. Specifically, for each algorithm we report the average runtime and the sample size.}
   \item (Optional) Anonymized source code, with specification of all dependencies, including external libraries. \textbf{Yes}
 \end{enumerate}

 \item For any theoretical claim, check if you include:
 \begin{enumerate}
   \item Statements of the full set of assumptions of all theoretical results. \textbf{Yes}
   \item Complete proofs of all theoretical results. \textbf{Yes}
   \item Clear explanations of any assumptions. \textbf{Yes}     
 \end{enumerate}

 \item For all figures and tables that present empirical results, check if you include:
 \begin{enumerate}
   \item The code, data, and instructions needed to reproduce the main experimental results (either in the supplemental material or as a URL). \textbf{Yes}
   \item All the training details (e.g., data splits, hyperparameters, how they were chosen). \textbf{Yes}
    \item A clear definition of the specific measure or statistics and error bars (e.g., with respect to the random seed after running experiments multiple times). \textbf{Yes}
    \item A description of the computing infrastructure used. (e.g., type of GPUs, internal cluster, or cloud provider). \textbf{Yes}
 \end{enumerate}

 \item If you are using existing assets (e.g., code, data, classifiers) or curating/releasing new assets, check if you include:
 \begin{enumerate}
   \item Citations of the creator If your work uses existing assets. \textbf{Yes}
   \item The license information of the assets, if applicable. \textbf{Yes}
   \item New assets either in the supplemental material or as a URL, if applicable. \textbf{Yes}
   \item Information about consent from data providers/curators. \textbf{Not Applicable}
   \item Discussion of sensible content if applicable, e.g., personally identifiable information or offensive content. \textbf{Not Applicable}
 \end{enumerate}

 \item If you used crowdsourcing or conducted research with human subjects, check if you include:
 \begin{enumerate}
   \item The full text of instructions given to participants and screenshots. \textbf{Not Applicable}
   \item Descriptions of potential participant risks, with links to Institutional Review Board (IRB) approvals if applicable. \textbf{Not Applicable}
   \item The estimated hourly wage paid to participants and the total amount spent on participant compensation. \textbf{Not Applicable}
 \end{enumerate}

 \end{enumerate}

\onecolumn

\appendix
\section*{Appendix}
\addcontentsline{toc}{section}{Appendices}
\renewcommand{\thesubsection}{\Alph{subsection}}
\section{Density Estimate Instances}
\label{densityestimates}
\begin{definition}
    A kernel density estimator (KDE) $\hat{q}_{\textnormal{KDE}}$ is defined as \begin{equation}
    \begin{aligned}
        \hat{q}_{\textnormal{KDE}}(\boldsymbol{x}) = \sum_{i=1}^m \boldsymbol{w}_i k(\boldsymbol{x}, \boldsymbol{x}_i),
    \end{aligned} \nonumber
\end{equation}
where $k(\cdot,\cdot)$ denotes a suitable kernel function, $\boldsymbol{x}_i$ denotes the $i-$th correctly classified sample in the training data set and $\boldsymbol{w}_i > 0$ denotes the weighting of the $i-$th sample.
\end{definition}
For the KDE plausibility term, we utilize a Gaussian normal kernel of bandwidth parameter $\sigma > 0$ (standard choice from \cite{racine2008nonparametric}) $k(\boldsymbol{x},\boldsymbol{x}_i):= e^{-\theta_2( \boldsymbol{x},\boldsymbol{x}_i)/{2\sigma^2}}$
 for correctly classified points $\boldsymbol{x}_i$ and we set $\boldsymbol{w}_i =1/m,$ for $i\in [m]$.
\begin{definition}
A Gaussian mixture classifier (GMM) $\hat{q}_{\textnormal{GMM}}$ with $m$ components is defined as
\begin{equation}
    \begin{aligned}
        \hat{q}_{\textnormal{GMM}}(\boldsymbol{x}) = \sum_{i=1}^m \pi_i \mathcal{N}(\boldsymbol{x} | \boldsymbol{\mu}_i, \boldsymbol{\Sigma}_i),
    \end{aligned} \nonumber
\end{equation}
where $\pi_i$ represents the prior probability of the $i-$th component, $\boldsymbol{\mu}_i$ and $\boldsymbol{\Sigma}_i$ denote the mean and covariance of the $i-$th component, a $d-$dimensional Gaussian density.
\end{definition}
\begin{definition}
\textnormal{(Density Gravity on an Instance (\citep{zhang2023density})}
\label{densitygravity}
    For $\boldsymbol{x} \in \mathcal{D},$ let $N_k(\boldsymbol{x}) = \{ \boldsymbol{x}_1,...,\boldsymbol{x}_k \}$ be the set of $k-$NNs of $\boldsymbol{x}$ in $\mathcal{D}$. The $k-$local density of $\boldsymbol{x}$ is defined by $\rho(\boldsymbol{x}) = |N_k(\boldsymbol{x})|(\sum_{\boldsymbol{x}_i \in N_k(\boldsymbol{x})} \|\boldsymbol{x} - \boldsymbol{x}_i\|_p)^{-1}.$ The set of relative local densities $\{ \hat{\rho}_1,..., \hat{\rho}_k \}$ of the points in $N_k(\boldsymbol{x})$ is the result of normalization as $\hat{\rho}_i=\frac{\rho_i}{\sum_{i=1}^k \rho_i}, \rho_i \geq 0, i \in [k]$. Then, the density gravity $G$ of $\boldsymbol{x}$ on $\mathcal{D}$ is defined as 
    \begin{equation}
    \begin{aligned}
        G(\boldsymbol{x}) = \sum_{i=1}^k \hat{\rho}_i \boldsymbol{x}_i, \ where \ \boldsymbol{x}_i \in N_k(\boldsymbol{x}), \ \sum_{i=1}^k \hat{\rho}_i = 1, \ and \ \hat{\rho}_i \geq 0.
    \end{aligned} \nonumber
    \end{equation}
\end{definition}
Note that the maximum operator $\textnormal{rd}_k(\boldsymbol{x},\cdot)$ of Definition \ref{LocalOutlierFactor} is linearized for simplicity. In simple words, the density gravity of $\boldsymbol{x}$, finds the closest point $G(\boldsymbol{x})$ that lies in a high-density data region by finding a convex combination of $k-$nearest neighbors of $\boldsymbol{x}$ weighted by their local density. Note that by definition, the local density, denoted by $\rho(\boldsymbol{x})$ in Definition \ref{densitygravity}, is higher for a point $\boldsymbol{x}$ if it has more neighbors that are closer to it. 
\section{Issues with Using Many Neighbors for the \texorpdfstring{$\textbf{S-CFE}_{\textbf{kNN}}$}{Lg} Approach} \label{manyneighborsissue}

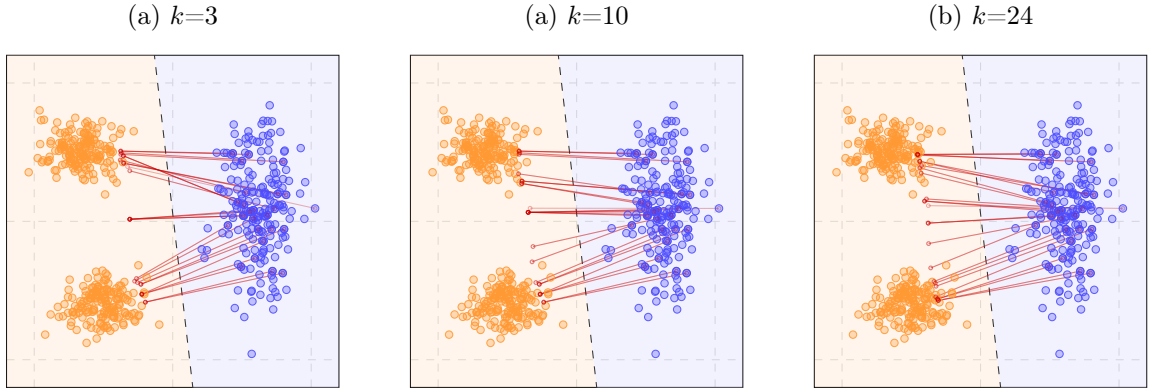
\begin{figure*}[ht]
    \centering
    \def\lims{12} 
\def\mscale{0.7} 
\def\wid{6cm}
\def\hig{6cm}

\pgfplotstableread[col sep = comma]{kNNclass1.csv}\classone
\pgfplotstableread[col sep = comma]{kNNclass0.csv}\classzero
\pgfplotstableread[col sep = comma]{perceptron.csv}\perceptron
\pgfplotstableread[col sep = comma]{kNN3.csv}\kNNtre
\pgfplotstableread[col sep = comma]{kNN10.csv}\kNNten
\pgfplotstableread[col sep = comma]{kNN24.csv}\kNNtwf

\pgfplotsset{select coords between index/.style 2 args={
    x filter/.code={
        \ifnum\coordindex<#1\def\pgfmathresult{}\fi
        \ifnum\coordindex>#2\def\pgfmathresult{}\fi
    }
}}
\begin{tikzpicture}
\begin{axis}[
        width=\wid, height=\hig,
        grid = major,
        grid style={dashed, gray!30},
        xmin=-\lims, xmax=\lims,
        ymin=-\lims, ymax=\lims,
        ticks=none,
        legend style={at={(0.245,0.97)}, anchor=north},
        title={(a) $k$=3}
     ]

    \path[name path=below] (axis cs:-\lims,-\lims) -- (axis cs:\lims,-\lims);
    \path[name path=above] (axis cs:-\lims,\lims) -- (axis cs:\lims,\lims);
    \addplot[only marks, mark options={scale=\mscale}, blue!70!white, fill opacity=0.3, draw opacity=0.9] table[x=x, y=y] {\classone};
    \addplot[only marks, mark options={scale=\mscale}, orange!80!white, fill opacity=0.3, draw opacity=0.9] table[x=x, y=y] {\classzero};
    \addplot[name path=DB, dashed, black!80!white] table[x expr=\thisrowno{0} / 10 * \lims, y expr=\thisrowno{1} / 10 * \lims ] {\perceptron};
     \addplot[fill=orange!30!white, nearly transparent] fill between[of=DB and below];
     \addplot[fill=blue!20!white, nearly transparent] fill between[of=DB and above];

    \pgfplotstablegetcolsof{\kNNtre}
    \foreach \idx in {0,...,\the\numexpr\pgfplotsretval-2\relax}{
        \addplot[mark=o, mark options={scale=0.5*\mscale}, red!80!black, fill opacity=0.3, draw opacity=0.3] table[x=x\the\numexpr\idx/2\relax, y=y\the\numexpr\idx/2\relax] {\kNNtre};
        \addplot[only marks, mark options={scale=\mscale}, blue!70!white, fill opacity=0.3, draw opacity=0.9, select coords between index={0}{0}] table[x=x\the\numexpr\idx/2\relax, y=y\the\numexpr\idx/2\relax] {\kNNtre};
    }
\end{axis}
\end{tikzpicture}
\hspace{20pt}
\begin{tikzpicture}
\begin{axis}[
        width=\wid, height=\hig,
        grid = major,
        grid style={dashed, gray!30},
        xmin=-\lims, xmax=\lims,
        ymin=-\lims, ymax=\lims,
        ticks=none,
        legend style={at={(0.245,0.97)}, anchor=north},
        title={(a) $k$=10}
     ]

    \path[name path=below] (axis cs:-\lims,-\lims) -- (axis cs:\lims,-\lims);
    \path[name path=above] (axis cs:-\lims,\lims) -- (axis cs:\lims,\lims);
    \addplot[only marks, mark options={scale=\mscale}, blue!70!white, fill opacity=0.3, draw opacity=0.9] table[x=x, y=y] {\classone};
    \addplot[only marks, mark options={scale=\mscale}, orange!80!white, fill opacity=0.3, draw opacity=0.9] table[x=x, y=y] {\classzero};
    \addplot[name path=DB, dashed, black!80!white] table[x expr=\thisrowno{0} / 10 * \lims, y expr=\thisrowno{1} / 10 * \lims ] {\perceptron};
     \addplot[fill=orange!30!white, nearly transparent] fill between[of=DB and below];
     \addplot[fill=blue!20!white, nearly transparent] fill between[of=DB and above];

    \pgfplotstablegetcolsof{\kNNten}
    \foreach \idx in {0,...,\the\numexpr\pgfplotsretval-2\relax}{
        \addplot[mark=o, mark options={scale=0.5*\mscale}, red!80!black, fill opacity=0.3, draw opacity=0.3] table[x=x\the\numexpr\idx/2\relax, y=y\the\numexpr\idx/2\relax] {\kNNten};
        \addplot[only marks, mark options={scale=\mscale}, blue!70!white, fill opacity=0.3, draw opacity=0.9, select coords between index={0}{0}] table[x=x\the\numexpr\idx/2\relax, y=y\the\numexpr\idx/2\relax] {\kNNten};
    }
\end{axis}
\end{tikzpicture}
\hspace{20pt}
\begin{tikzpicture}
\begin{axis}[
        width=\wid, height=\hig,
        grid = major,
        grid style={dashed, gray!30},
        xmin=-\lims, xmax=\lims,
        ymin=-\lims, ymax=\lims,
        ticks=none,
        legend style={at={(0.245,0.97)}, anchor=north},
        title={(b) $k$=24}
     ]

    \path[name path=below] (axis cs:-\lims,-\lims) -- (axis cs:\lims,-\lims);
    \path[name path=above] (axis cs:-\lims,\lims) -- (axis cs:\lims,\lims);
    \addplot[only marks, mark options={scale=\mscale}, blue!70!white, fill opacity=0.3, draw opacity=0.9] table[x=x, y=y] {\classone};
    \addplot[only marks, mark options={scale=\mscale}, orange!80!white, fill opacity=0.3, draw opacity=0.9] table[x=x, y=y] {\classzero};
    \addplot[name path=DB, dashed, black!80!white] table[x expr=\thisrowno{0} / 10 * \lims, y expr=\thisrowno{1} / 10 * \lims ] {\perceptron};
     \addplot[fill=orange!30!white, nearly transparent] fill between[of=DB and below];
     \addplot[fill=blue!20!white, nearly transparent] fill between[of=DB and above];

    \pgfplotstablegetcolsof{\kNNtwf}
    \foreach \idx in {0,...,\the\numexpr\pgfplotsretval-2\relax}{
        \addplot[mark=o, mark options={scale=0.5*\mscale}, red!80!black, fill opacity=0.3, draw opacity=0.3] table[x=x\the\numexpr\idx/2\relax, y=y\the\numexpr\idx/2\relax] {\kNNtwf};
        \addplot[only marks, mark options={scale=\mscale}, blue!70!white, fill opacity=0.3, draw opacity=0.9, select coords between index={0}{0}] table[x=x\the\numexpr\idx/2\relax, y=y\the\numexpr\idx/2\relax] {\kNNtwf};
    }
\end{axis}
\end{tikzpicture}
\vspace{15pt}
    \caption{A toy example illustrating the positioning of convex combinations of $k-$NNs obtained via density gravity relative to the original points.}
    \label{differentneighbors}
\end{figure*}

In \cref{differentneighbors} we experiment with choosing different neighbors for the plausibility term given by density gravity, which minimizes the distance between the CFE and a point in the convex combination of the $k-$nearest neighbors of given factual data point, weighted by their local density. Choosing higher values of $k$ results in neighbors being chosen from different high-density areas of correctly classified data points, thus the density gravity point can result in low-density areas (as can be seen in the middle and the right figure in \cref{differentneighbors}).

\section{Our S-CFE Approach vs MIP-LIVE-m=1}
\label{ourvsmiplive}

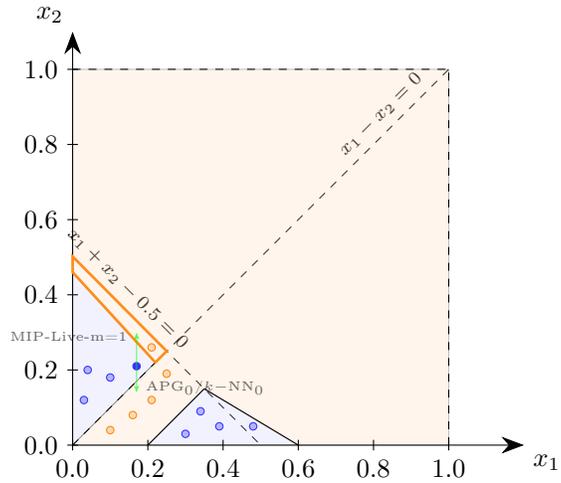
\begin{figure}[H]
\vspace{.3in}
\centering
\begin{tikzpicture}
\def\mscale{0.7} 
    \draw[thin,-{Stealth[length=3mm]}] (0,0) -- (6,0) node[anchor=north west] {$x_1$};
\draw[thin,-{Stealth[length=3mm]}] (0,0) -- (0,5.5) node[anchor= south east] {$x_2$};
\draw[dashed] (5,0) -- (5,5);
\draw[dashed] (0,5) -- (5,5);
\draw[dashed] (0,0) -- (5,5) node[pos=0.85,sloped,above] {\scriptsize $x_1-x_2=0$};
\draw[dashed] (0,2.5) -- (2.5,0) node[pos=0.23,sloped,above] {\scriptsize $x_1+x_2-0.5=0$};
\draw (1,0) -- (1.75,0.75);
\draw (3,0) -- (1.75,0.75);
\draw (0,2.3) -- (1.1,1.1);

\draw[fill=blue!20!white, nearly transparent]  (1,0) -- (1.75,0.75) -- (3,0) -- cycle;
\draw[fill=blue!20!white, nearly transparent]  (0.0,0.0) -- (0.0,2.3) -- (1.1,1.1) -- cycle;
\draw[fill=orange!30!white, nearly transparent]  (0,0) -- (1.1,1.1) -- (0,2.3) -- (0,5) -- (5,5) -- (5,0) -- (3,0) -- (1.75, 0.75) -- (1,0)  -- cycle;

\draw[line width=0.35mm, orange!90!white] (1.1,1.1) -- (1.25,1.25) -- (0,2.51) -- (0,2.3) -- cycle;

\draw[blue!80!white, fill=blue!30!white] (0.15,0.6) circle (\mscale*2pt);
\draw[blue!80!white, fill=blue!30!white] (0.2,1) circle (\mscale*2pt);
\draw[blue!80!white, fill=blue!30!white] (0.5,0.9) circle (\mscale*2pt);
\draw[blue!80!white, fill=blue!80!white] (0.85,1.05) circle (\mscale*2pt);

\draw[green, opacity=0.6, thin,-{Stealth[length=1mm]}] (0.85,1.05) -- (0.85,1.5) node[pos=0.9,left] {\tiny {\color{black} MIP-Live-m=1}};
\draw[green, opacity=0.6, thin,-{Stealth[length=1mm]}] (0.85,1.05) -- (0.85,0.7) node[pos=0.8,right] {\tiny {\color{black} S-CFE}};

\draw[orange!95!white, fill=orange!30!white] (0.5,0.2) circle (\mscale*2pt);
\draw[orange!95!white, fill=orange!30!white] (0.8,0.4) circle (\mscale*2pt);
\draw[orange!95!white, fill=orange!30!white] (1.05,0.6) circle (\mscale*2pt);
\draw[orange!95!white, fill=orange!30!white] (1.25,0.95) circle (\mscale*2pt);
\draw[orange!95!white, fill=orange!30!white] (1.05,1.3) circle (\mscale*2pt);



\draw[blue!80!white, fill=blue!30!white] (1.5,0.15) circle (\mscale*2pt);
\draw[blue!80!white, fill=blue!30!white] (1.7,0.45) circle (\mscale*2pt);
\draw[blue!80!white, fill=blue!30!white] (1.95,0.25) circle (\mscale*2pt);
\draw[blue!80!white, fill=blue!30!white] (2.4,0.25) circle (\mscale*2pt);

\foreach \x in {0.0,0.2,0.4,0.6,0.8,1.0}
    \draw (5*\x cm,2pt) -- (5*\x cm,-2pt) node[anchor=north] {$\x$};
\foreach \y in {0.0,0.2,0.4,0.6,0.8,1.0}
    \draw (2pt, 5*\y cm) -- (-2pt,5*\y cm) node[anchor=east] {$\y$};
\end{tikzpicture}
\vspace{.3in}
\caption{Example reproduced from \cite{tsiourvas2024manifold}. Compares geometrically MIP-Live-m=1 \citep{tsiourvas2024manifold} vs. our S-CFE approach. The generated CFE of our method resides in a high-density region and is sparse. MIP-Live-m=1 considerably restricts the working space - the small bounded red region, and uses only 1 neighbor for the LOF manifold adhering constraint.}
\label{miplivevsours}
\end{figure}

\section{Additional Experiments}
\label{additionalexperiments}
We provide additional results in \cref{tab:linearextended} and \cref{tab:nonlinearextended}, demonstrating that our method generates sparse, manifold-aligned CFEs that closely follow the data manifold while maintaining efficient computation times. These findings align with \cref{quanteval}. 

We exclude CEM \citep{dhurandhar2018explanations} from further experiments since \cite{zhang2023density} already benchmark against it. Instead, we present additional results using the closed-form solution for the proximal operator in \cref{quadraticapproxprox} for $p=1/2$ \citep{cao2013fast, lin2024computing}. The solution, for $i \in [d]$ and the inverse Lipschitz constant $L$ replaced by a step size sequence $(\sigma_t)_{t \in \mathbb{N}}$, is given by
\begin{align*}
\begin{gathered}
[\boldsymbol{x}_{cf, \frac{1}{2}}^{t+1}]_i = \begin{cases}
         \frac{2}{3}[S_{\sigma_t} (\boldsymbol{x}^t,y_{cf})]_i \biggl( 1+\cos\Bigl(\frac{2\pi}{3} -\frac{2\phi_{2\beta \sigma_t}([S_{\sigma_t} (\boldsymbol{x}^t,y_{cf})]_i)}{3}\Bigr) \biggr), 
         \ & \text{if}\  |[S_{\sigma_t} (\boldsymbol{x}^t,y_{cf})]_i| > g(2\beta \sigma_t),\\
         0, \ \ & \text{otherwise,}
     \end{cases} \nonumber 
    \end{gathered} 
\end{align*}
where $\phi_{2\beta \sigma_t}([S_{\sigma_t} (\boldsymbol{x}^t,y_{cf})]_i) = \arccos\left( \frac{ \beta \sigma_t}{4}\left( \frac{|[S_{\sigma_t} (\boldsymbol{x}^t,y_{cf})]_i|}{3} \right)^{-\frac{3}{2}} \right),$ and $g(2\beta \sigma_t)=\frac{\sqrt[3]{54}}{4}(2\beta \sigma_t)^{\frac{2}{3}}$.

\begin{figure*}[ht]
\centering 
    \footnotesize
    \definecolor{plotcolor1}{HTML}{1b9e77}
\definecolor{plotcolor2}{HTML}{d95f02}
\definecolor{plotcolor3}{HTML}{7570b3}
\definecolor{plotcolor4}{HTML}{e7298a}

    \begin{tikzpicture}
    \pgftransformscale{1}
    \begin{groupplot}[group style={group size= 3 by 3, horizontal sep=1cm, vertical sep=1.3cm},height=5cm,width=6cm, , legend columns=-1]
        \nextgroupplot[
        y label style={at={(axis description cs:0.14,0.5)},anchor=south},
        ylabel = {$\ell_2-$norm output},
        xmajorgrids=true,
        ymajorgrids=true,
        every x tick scale label/.style={at={(xticklabel cs:0.5)}, color=white},
        title={Boston Housing, $p=0$},
        legend style={{at=(0.025,0.025)}, anchor=south west},
        legend to name=legend_plt,
        cycle multiindex list={plotcolor1, plotcolor2, plotcolor3, plotcolor4, plotcolor5, plotcolor6, plotcolor7\nextlist mark list}]
        
        \addplot
            table[x=l2,y=l2_no_plaus,col sep=comma]{BostonRobustness0.csv};
        \addlegendentry{No Plaus.};
        \addplot
            table[x=l2,y=l2_kde,col sep=comma]{BostonRobustness0.csv};
        \addlegendentry{KDE};
        \addplot
            table[x=l2,y=l2_gmm,col sep=comma]{BostonRobustness0.csv};
        \addlegendentry{GMM};
        \addplot
            table[x=l2,y=l2_knn,col sep=comma]{BostonRobustness0.csv};
        \addlegendentry{$k$-NN};
        
        \nextgroupplot[
        xmajorgrids=true,
        ymajorgrids=true,
        every x tick scale label/.style={at={(xticklabel cs:0.5)}, color=white},
        title={Boston Housing, $p=\frac{1}{2}$},
        legend style={{at=(0.025,0.025)}, anchor=south west},
        legend to name=legend_plt,
        cycle multiindex list={plotcolor1, plotcolor2, plotcolor3, plotcolor4, plotcolor5, plotcolor6, plotcolor7\nextlist mark list}]
        
        \addplot
            table[x=l2,y=l2_no_plaus,col sep=comma]{BostonRobustness1_2.csv};
        \addlegendentry{No Plaus.};
        \addplot
            table[x=l2,y=l2_kde,col sep=comma]{BostonRobustness1_2.csv};
        \addlegendentry{KDE};
        \addplot
            table[x=l2,y=l2_gmm,col sep=comma]{BostonRobustness1_2.csv};
        \addlegendentry{GMM};
        \addplot
            table[x=l2,y=l2_knn,col sep=comma]{BostonRobustness1_2.csv};
        \addlegendentry{$k$-NN};

        \nextgroupplot[
        xmajorgrids=true,
        ymajorgrids=true,
        every x tick scale label/.style={at={(xticklabel cs:0.5)}, color=white},
        title={Boston Housing, $p=1$},
        legend style={{at=(0.025,0.025)}, anchor=south west},
        legend to name=legend_plt,
        cycle multiindex list={plotcolor1, plotcolor2, plotcolor3, plotcolor4, plotcolor5, plotcolor6, plotcolor7\nextlist mark list}]
        
        \addplot
            table[x=l2,y=l2_no_plaus,col sep=comma]{BostonRobustness1.csv};
        \addlegendentry{No Plaus.};
        \addplot
            table[x=l2,y=l2_kde,col sep=comma]{BostonRobustness1.csv};
        \addlegendentry{KDE};
        \addplot
            table[x=l2,y=l2_gmm,col sep=comma]{BostonRobustness1.csv};
        \addlegendentry{GMM};
        \addplot
            table[x=l2,y=l2_knn,col sep=comma]{BostonRobustness1.csv};
        \addlegendentry{$k$-NN};

        \nextgroupplot[
        y label style={at={(axis description cs:0.14,0.5)},anchor=south},
        ylabel = {$\ell_2-$norm output},
        xmajorgrids=true,
        ymajorgrids=true,
        every x tick scale label/.style={at={(xticklabel cs:0.5)}, color=white},
        title={Wine, $p=0$},
        legend style={{at=(0.025,0.025)}, anchor=south west},
        legend to name=legend_plt,
        cycle multiindex list={plotcolor1, plotcolor2, plotcolor3, plotcolor4, plotcolor5, plotcolor6, plotcolor7\nextlist mark list}]
        
        \addplot
            table[x=l2,y=l2_no_plaus,col sep=comma]{WineRobustness0.csv};
        \addlegendentry{No Plaus.};
        \addplot
            table[x=l2,y=l2_kde,col sep=comma]{WineRobustness0.csv};
        \addlegendentry{KDE};
        \addplot
            table[x=l2,y=l2_gmm,col sep=comma]{WineRobustness0.csv};
        \addlegendentry{GMM};
        \addplot
            table[x=l2,y=l2_knn,col sep=comma]{WineRobustness0.csv};
        \addlegendentry{$k$-NN};
        \coordinate (top) at (rel axis cs:0,1);
        
        \nextgroupplot[
        xmajorgrids=true,
        ymajorgrids=true,
        every x tick scale label/.style={at={(xticklabel cs:0.5)}, color=white},
        title={Wine, $p=\frac{1}{2}$},
        legend style={{at=(0.025,0.025)}, anchor=south west},
        legend to name=legend_plt,
        cycle multiindex list={plotcolor1, plotcolor2, plotcolor3, plotcolor4, plotcolor5, plotcolor6, plotcolor7\nextlist mark list}]
        
        \addplot
            table[x=l2,y=l2_no_plaus,col sep=comma]{WineRobustness1_2.csv};
        \addlegendentry{No Plaus.};
        \addplot
            table[x=l2,y=l2_kde,col sep=comma]{WineRobustness1_2.csv};
        \addlegendentry{KDE};
        \addplot
            table[x=l2,y=l2_gmm,col sep=comma]{WineRobustness1_2.csv};
        \addlegendentry{GMM};
        \addplot
            table[x=l2,y=l2_knn,col sep=comma]{WineRobustness1_2.csv};
        \addlegendentry{$k$-NN};

        \nextgroupplot[
        xmajorgrids=true,
        ymajorgrids=true,
        every x tick scale label/.style={at={(xticklabel cs:0.5)}, color=white},
        title={Wine, $p=1$},
        legend style={{at=(0.025,0.025)}, anchor=south west},
        legend to name=legend_plt,
        cycle multiindex list={plotcolor1, plotcolor2, plotcolor3, plotcolor4, plotcolor5, plotcolor6, plotcolor7\nextlist mark list}]
        
        \addplot
            table[x=l2,y=l2_no_plaus,col sep=comma]{WineRobustness1.csv};
        \addlegendentry{No Plaus.};
        \addplot
            table[x=l2,y=l2_kde,col sep=comma]{WineRobustness1.csv};
        \addlegendentry{KDE};
        \addplot
            table[x=l2,y=l2_gmm,col sep=comma]{WineRobustness1.csv};
        \addlegendentry{GMM};
        \addplot
            table[x=l2,y=l2_knn,col sep=comma]{WineRobustness1.csv};
        \addlegendentry{$k$-NN};

        \nextgroupplot[
        x label style={at={(axis description cs:0.5,0.0)},anchor=north},
        y label style={at={(axis description cs:0.14,0.5)},anchor=south},
        ylabel = {$\ell_2-$norm output},
        xlabel = {$\ell_2-$norm input},
        xmajorgrids=true,
        ymajorgrids=true,
        every x tick scale label/.style={at={(xticklabel cs:0.5)}, color=white},
        title={Wisconsin, $p=0$},
        legend style={{at=(0.025,0.025)}, anchor=south west},
        legend to name=legend_plt,
        cycle multiindex list={plotcolor1, plotcolor2, plotcolor3, plotcolor4, plotcolor5, plotcolor6, plotcolor7\nextlist mark list}]
        
        \addplot
            table[x=l2,y=l2_no_plaus,col sep=comma]{WisconsinRobustness0.csv};
        \addlegendentry{No Plaus.};
        \addplot
            table[x=l2,y=l2_kde,col sep=comma]{WisconsinRobustness0.csv};
        \addlegendentry{KDE};
        \addplot
            table[x=l2,y=l2_gmm,col sep=comma]{WisconsinRobustness0.csv};
        \addlegendentry{GMM};
        \addplot
            table[x=l2,y=l2_knn,col sep=comma]{WisconsinRobustness0.csv};
        \addlegendentry{$k$-NN};
        
        \nextgroupplot[
        x label style={at={(axis description cs:0.5,0.0)},anchor=north},
        xlabel = {$\ell_2-$norm input},
        xmajorgrids=true,
        ymajorgrids=true,
        every x tick scale label/.style={at={(xticklabel cs:0.5)}, color=white},
        title={Wisconsin, $p=\frac{1}{2}$},
        legend style={{at=(0.025,0.025)}, anchor=south west},
        legend to name=legend_plt,
        cycle multiindex list={plotcolor1, plotcolor2, plotcolor3, plotcolor4, plotcolor5, plotcolor6, plotcolor7\nextlist mark list}]
        
        \addplot
            table[x=l2,y=l2_no_plaus,col sep=comma]{WisconsinRobustness1_2.csv};
        \addlegendentry{No Plaus.};
        \addplot
            table[x=l2,y=l2_kde,col sep=comma]{WisconsinRobustness1_2.csv};
        \addlegendentry{KDE};
        \addplot
            table[x=l2,y=l2_gmm,col sep=comma]{WisconsinRobustness1_2.csv};
        \addlegendentry{GMM};
        \addplot
            table[x=l2,y=l2_knn,col sep=comma]{WisconsinRobustness1_2.csv};
        \addlegendentry{$k$-NN};

        \nextgroupplot[
        x label style={at={(axis description cs:0.5,0.0)},anchor=north},
        xlabel = {$\ell_2-$norm input},
        xmajorgrids=true,
        ymajorgrids=true,
        every x tick scale label/.style={at={(xticklabel cs:0.5)}, color=white},
        title={Wisconsin, $p=1$},
        legend style={{at=(0.025,0.025)}, anchor=south west},
        legend to name=legend_plt,
        cycle multiindex list={plotcolor1, plotcolor2, plotcolor3, plotcolor4, plotcolor5, plotcolor6, plotcolor7\nextlist mark list}]
        
        \addplot
            table[x=l2,y=l2_no_plaus,col sep=comma]{WisconsinRobustness1.csv};
        \addlegendentry{No Plaus.};
        \addplot
            table[x=l2,y=l2_kde,col sep=comma]{WisconsinRobustness1.csv};
        \addlegendentry{KDE};
        \addplot
            table[x=l2,y=l2_gmm,col sep=comma]{WisconsinRobustness1.csv};
        \addlegendentry{GMM};
        \addplot
            table[x=l2,y=l2_knn,col sep=comma]{WisconsinRobustness1.csv};
        \addlegendentry{$k$-NN};
        
        \coordinate (bot) at (rel axis cs:0.815,0);
    \end{groupplot}
    \path (top)--(bot) coordinate[midway] (group center);
    \node[inner sep=0pt] at ([yshift=-0.5cm, xshift=0.3cm] group center |- current bounding box.south) {\pgfplotslegendfromname{legend_plt}};
    \normalsize
    \end{tikzpicture}
    \vspace{15pt}
\caption{Robustness of the different methods. The distance of the input data points to the original data points on the $x$-axis and the distance of the generated CFEs to the CFE generated from the original data points on the $y$-axis. Tested on 100 data points from each data set.}
\label{robustnessplots}
\vspace{-5pt}
\end{figure*}
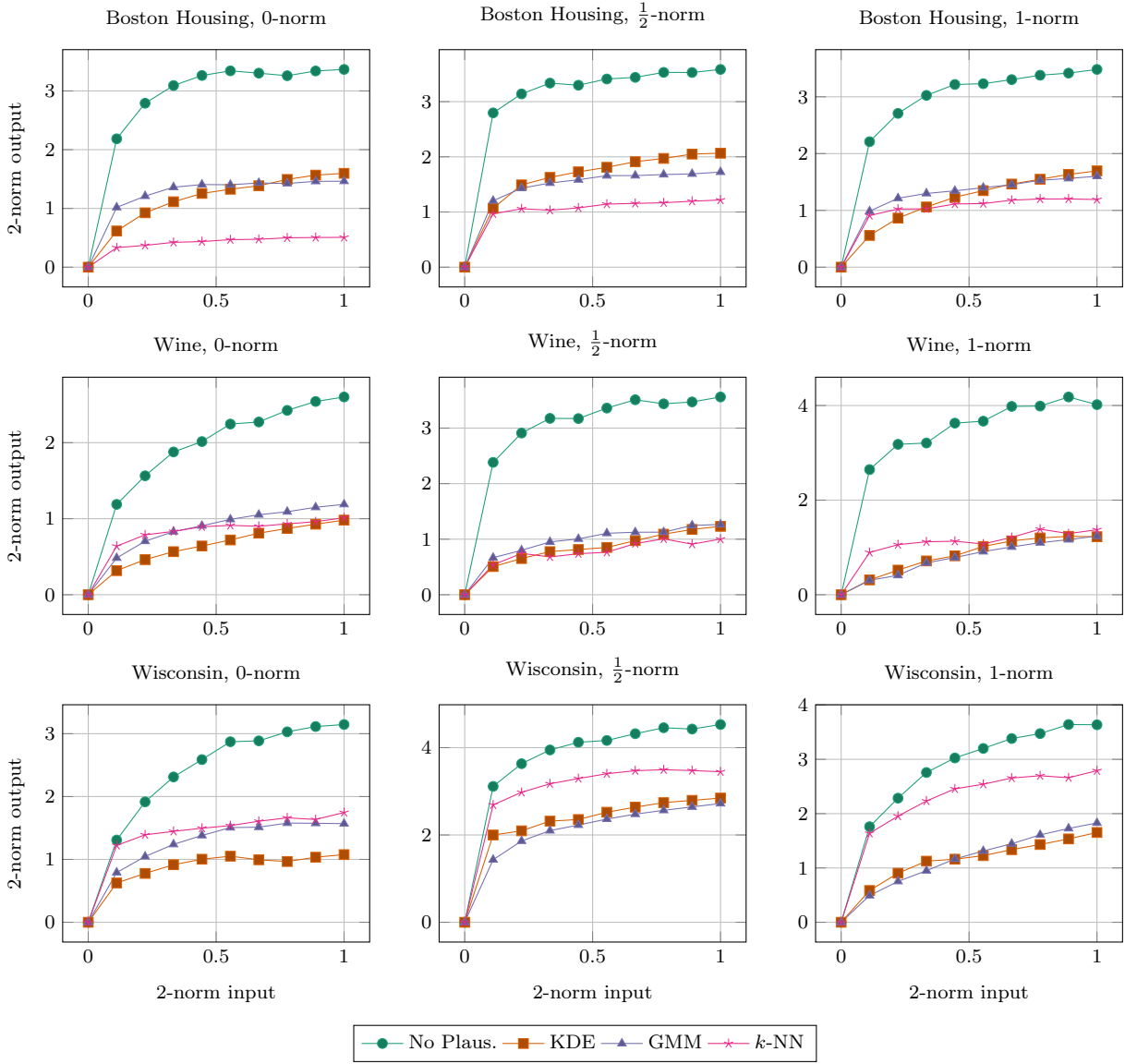

Results in \cref{robustnessplots} of \cref{extendedrobustness} further confirm findings of section \cref{robustness} that our S-CFE generated CFEs are robust to input shifts.

\begin{table*}
    \caption{CFEs for linear classifiers on the Boston Housing and Wine datasets. The dimensionality of both datasets has been reduced to 8 using PCA. Evaluated 100 test points. The S-CFE $p=0$ variants use sparsity constraints, while the other variants use the corresponding norm as a regularizer. The compute time is given in seconds per 100 CFEs. The compute time is given in seconds per 100 CFEs. We denote by DCFE and PCFE the methods proposed by \cite{zhang2023density} and \cite{artelt2020convex}, respectively.}
    \centering
    \begin{tabular}{ccccccccc}
    \toprule
        Dataset & Method & Validity & $\ell_2$ & $\ell_0$ & LOF & KDE & GMM & Time\\
    \midrule
        \multirow{10}{*}{\begin{tabular}{c}Wine\\8 features\\PCA\end{tabular}}& S-CFE$_{KDE}$ $p=0$ & 100 & 2.10 & \textbf{1.00} & \textbf{0.98} & \textbf{-2.61} & -19.0 & 9.88 \\
        & S-CFE$_{KDE}$ $p=\frac{1}{2}$ & 100 & 1.93 & 1.52 & \textbf{0.98} & -2.71 & -19.1 & 12.6 \\
        & S-CFE$_{KDE}$ $p=1$ & 100 & 2.11 & 2.02 & 0.99 & -2.68 & -23.5 & 11.0 \\
    \cmidrule(lr){2-9}
        & S-CFE$_{GMM}$ $p=0$ & 100 & 2.09 & \textbf{1.00} & \textbf{0.98} & -2.86 & \textbf{-14.3} & 11.4 \\
        & S-CFE$_{GMM}$ $p=\frac{1}{2}$ & 100 & 2.86 & 1.31 & 0.99 & -2.89 & -15.3 & 13.6 \\
        & S-CFE$_{GMM}$ $p=1$ & 100 & 2.40 & 2.00 & 0.99 & -3.01 & -20.1 & 11.6 \\
    \cmidrule(lr){2-9}
        & S-CFE$_{k-\text{NN}}$ $p=0$ & 100 & 2.15 & \textbf{1.00} & 0.99 & -2.92 & -19.8 & 4.50 \\
        & S-CFE$_{k-\text{NN}}$ $p=\frac{1}{2}$ & 100 & 2.49 & 1.55 & 1.01 & 2.93 & -20.9 & 4.79 \\
        & DCFE & 100 & 2.08 & 1.99 & 1.01 & -2.96 & -23.4 & \textbf{3.49} \\
    \cmidrule(lr){2-9}
        & PCFE & 100 & \textbf{1.39} & 1.37 & 0.99 & -2.93 & -20.6 & 22.3 \\
    \midrule
    \midrule
        \multirow{10}{*}{\begin{tabular}{c}Housing\\8 features\\PCA\end{tabular}}& S-CFE$_{KDE}$ $p=0$ & 100 & 3.06 & \textbf{1.00} & 1.22 & \textbf{-2.55} & -14.1 & 11.8 \\
        & S-CFE$_{KDE}$ $p=\frac{1}{2}$ & 100 & 2.82 & 1.31 & \textbf{1.20} & -2.69 & -15.8 & 13.2 \\
        & S-CFE$_{KDE}$ $p=1$ & 100 & 2.14 & 2.44 & 1.23 & -2.75 & -14.8 & 10.3 \\
    \cmidrule(lr){2-9}
        & S-CFE$_{GMM}$ $p=0$ & 100 & 2.62 & \textbf{1.00} & \textbf{1.20} & -2.89 & \textbf{-10.8} & 12.1 \\
        & S-CFE$_{GMM}$ $p=\frac{1}{2}$ & 100 & 3.01 & 1.92 & 1.23 & -2.84 & -11.3 & 12.5 \\ 
        & S-CFE$_{GMM}$ $p=1$ & 100 & 2.73 & 3.19 & 1.26 & -2.92 & -11.6 & 12.0 \\
    \cmidrule(lr){2-9}
        & S-CFE$_{k-\text{NN}}$ $p=0$ & 100 & 2.51 & \textbf{1.00} & 1.34 & -3.01 & -14.2 & 4.61 \\
        & S-CFE$_{k-\text{NN}}$ $p=\frac{1}{2}$ & 100 & 3.05 & 1.87 & 1.38 & -3.01 & -20.8 & 4.48 \\
        & DCFE & 100 & 3.05 & 3.22 & 1.31 & -2.99 & -16.6 & \textbf{4.17} \\
    \cmidrule(lr){2-9}
        & PCFE & 100 & \textbf{1.61} & 1.13 & 1.31 & -3.06 & -11.4 & 26.7 \\
    \bottomrule
    \end{tabular}
    
    \label{tab:linearextended}
\end{table*}

\begin{table*}
   \caption{CFEs for DNN classifiers on the Boston Housing Wine datasets, and for a CNN classifier on the MNIST dataset. Evaluated on 1000 test points for MNIST and 100 test points for the other two datasets. The S-CFE $p=0$ variants use sparsity constraints, while the other variants use the corresponding norm as a regularizer. The compute time is given in seconds per 100 CFEs. We denote by DCFE the method proposed by \cite{zhang2023density}.}
    \centering
    \small
    \begin{tabular}{ccccccccc}
    \toprule
        Dataset & Method & Validity & $\ell_2$ & $\ell_0$ & LOF & KDE & GMM & Time\\
    \midrule
        \multirow{10}{*}{\begin{tabular}{c}Housing\\12 features\end{tabular}}& S-CFE$_{KDE}$ $p=0$ & 100 & 2.59 & \textbf{2.00} & 1.23 & -2.91 & -15.3 & 12.7 \\
        & S-CFE$_{KDE}$ $p=\frac{1}{2}$ & 100 & \textbf{2.41} & 2.78 & 1.18 & \textbf{-2.87} & -16.8 & 14.5 \\
        & S-CFE$_{KDE}$ $p=1$ & 100 & 2.52 & 6.30 & 1.27 & -3.02 & -17.7 & 12.3 \\
    \cmidrule(lr){2-9}
        & S-CFE$_{GMM}$ $p=0$ & 100 & 2.91 & \textbf{2.00} & \textbf{1.12} & -3.23 & \textbf{-12.8} & 13.3 \\
        & S-CFE$_{GMM}$ $p=\frac{1}{2}$ & 100 & 2.74 & 3.09 & 1.17 & -3.55 & -13.6 & 15.1 \\
        & S-CFE$_{GMM}$ $p=1$ & 100 & 2.76 & 5.76 & 1.24 & -3.47 & -14.1 & 12.6 \\
    \cmidrule(lr){2-9}
        & S-CFE$_{k-\text{NN}}$ $p=0$ & 100 & 3.64 & \textbf{2.00} & 1.17 & -3.08 & -17.6 & 5.85 \\
        & S-CFE$_{k-\text{NN}}$ $p=\frac{1}{2}$ & 100 & 3.61 & 4.04 & 1.20 & -3.31 & -19.1 & 6.04 \\
        & DCFE & 100 & 3.50 & 6.86 & 1.27 & -3.49 & -20.7 & \textbf{5.33} \\
    \midrule
    \midrule
        \multirow{10}{*}{\begin{tabular}{c}Wine\\13 features\end{tabular}}& S-CFE$_{KDE}$ $p=0$ & 100 & 3.31 & \textbf{2.00} & 0.99 & -2.87 & -24.1 & 12.4 \\
        & S-CFE$_{KDE}$ $p=\frac{1}{2}$ & 100 & 3.30 & 5.40 & \textbf{0.98} & \textbf{-2.76} & -21.8 & 11.8 \\
        & S-CFE$_{KDE}$ $p=1$ & 100 & \textbf{2.83} & 6.73 & 1.00 & -2.85 & -25.4 & 12.3 \\
    \cmidrule(lr){2-9}
        & S-CFE$_{GMM}$ $p=0$ & 100 & 3.44 & \textbf{2.00} & \textbf{0.98} & -3.11 & \textbf{-14.8} & 13.1 \\
        & S-CFE$_{GMM}$ $p=\frac{1}{2}$ & 100 & 3.83 & 5.73 & 0.99 & -2.98 & -15.9 & 12.6 \\
        & S-CFE$_{GMM}$ $p=1$ & 100 & 3.08 & 6.98 & 1.01 & -3.05 & -16.2 & 13.1 \\
    \cmidrule(lr){2-9}
        & S-CFE$_{k-\text{NN}}$ $p=0$ & 100 & 4.04 & \textbf{2.00} & 1.01 & -3.17 & -37.9 & 5.80 \\
        & S-CFE$_{k-\text{NN}}$ $p=\frac{1}{2}$ & 100 & 3.71 & 6.82 & 1.02 & -3.66 & -40.2 & 5.72 \\
        & DCFE & 100 & 3.21 & 7.13 & 1.03 & -3.77 & -42.1 & \textbf{4.95} \\
    \midrule
    \midrule
        \multirow{7}{*}{\begin{tabular}{c}MNIST\\784 features\end{tabular}}& S-CFE$_{GMM}$ $p=0$ & 99.1 & \textbf{6.74} & \textbf{25.0} & \textbf{1.21} & - & \textbf{-1058} & 55.3 \\
        & S-CFE$_{GMM}$ $p=\frac{1}{2}$ & 98.4 & 7.12 & 77.4 & 1.34 & - & -1112 & 57.3 \\
        & S-CFE$_{GMM}$ $p=1$ & 99.3 & 8.07 & 115 & 1.47 & - & -1132 & 54.2 \\
    \cmidrule(lr){2-9}
        & S-CFE$_{k-\text{NN}}$ $p=0$  & 99.8 & 7.04 & \textbf{25.0} & 1.30 & - & -1075 & 13.1 \\
        & S-CFE$_{k-\text{NN}}$ $p=\frac{1}{2}$ & \textbf{99.9} & 8.13 & 80.9 & 1.20 & - & -1069 & 12.7 \\
        & DCFE & 99.3 & 8.06 & 118 & 1.32 & - & -1122 & \textbf{11.8} \\
    \bottomrule
    \end{tabular}
    
    \label{tab:nonlinearextended}
\end{table*}

\end{document}